\documentclass[runningheads]{llncs}
\usepackage{graphicx}
\usepackage{comment}
\usepackage{amsmath,amssymb} 
\usepackage{color}

\usepackage{siunitx}
\usepackage{cite}
\usepackage{ctable}
\usepackage{hhline}
\usepackage{booktabs}
\usepackage{csquotes}
\usepackage{pifont}
\usepackage{multirow}
\usepackage{booktabs}
\usepackage{pifont}
\usepackage{mathtools}
\usepackage{subcaption}
\usepackage{hyperref}
\usepackage[super]{nth}

\newcolumntype{L}[1]{>{\raggedright\let\newline\\\arraybackslash\hspace{0pt}}m{#1}}
\newcolumntype{C}[1]{>{\centering\let\newline\\\arraybackslash\hspace{0pt}}m{#1}}
\newcolumntype{R}[1]{>{\raggedleft\let\newline\\\arraybackslash\hspace{0pt}}m{#1}}

\begin{document}
\pagestyle{headings}
\mainmatter

\title{DeepHandMesh: A Weakly-supervised \\ Deep Encoder-Decoder Framework \\ for High-fidelity Hand Mesh Modeling}

\titlerunning{DeepHandMesh}
%
\author{Gyeongsik Moon \inst{1} \and
Takaaki Shiratori \inst{2} \and
Kyoung Mu Lee \inst{1}}
\authorrunning{G. Moon et al.}
%
\institute{ECE \& ASRI, Seoul National University, Korea \and
Facebook Reality Labs \\
\email{\{mks0601,kyoungmu\}@snu.ac.kr}, \email{tshiratori@fb.com}}

\maketitle

\setcounter{footnote}{0} 

\begin{abstract}
Human hands play a central role in interacting with other people and objects. For realistic replication of such hand motions, high-fidelity hand meshes have to be reconstructed. In this study, we firstly propose DeepHandMesh, a weakly-supervised deep encoder-decoder framework for high-fidelity hand mesh modeling. We design our system to be trained in an end-to-end and weakly-supervised manner; therefore, it does not require groundtruth meshes. Instead, it relies on weaker supervisions such as 3D joint coordinates and multi-view depth maps, which are easier to get than groundtruth meshes and do not dependent on the mesh topology. Although the proposed DeepHandMesh is trained in a weakly-supervised way, it provides significantly more realistic hand mesh than previous fully-supervised hand models. Our newly introduced penetration avoidance loss further improves results by replicating physical interaction between hand parts. Finally, we demonstrate that our system can also be applied successfully to the 3D hand mesh estimation from general images.
Our hand model, dataset, and codes are publicly available\footnote{\url{https://mks0601.github.io/DeepHandMesh/}}.

\end{abstract}

\section{Introduction}

Social interactions are vital to humans: every day, we spend a large amount of time on interactions and communications with other people. While facial motion and speech play a central role in communication, important non-verbal information is also communicated via body motion, especially hand and finger motion, to emphasize our speech, clarify our ideas, and convey emotions. Modeling and replicating detailed hand geometry and motion is essential to enrich experience in various applications, including remote communications in virtual/augmented reality and digital storytelling such as movies and video games.

A pioneering work of hand geometry modeling is MANO by Romero~et al.~\cite{romero2017embodied}, which consists of linear models of identity- and pose-dependent correctives with linear blend skinning~(LBS) as an underlying mesh deformation algorithm. The model is learned in a fully-supervised manner by minimizing the per-vertex distance between output and groundtruth meshes that are obtained by registering a template mesh to 3D hand scans~\cite{bogo2014faust,hirshberg2012coregistration}.

\begin{figure}[t]
\begin{center}
\includegraphics[width=0.8\linewidth]{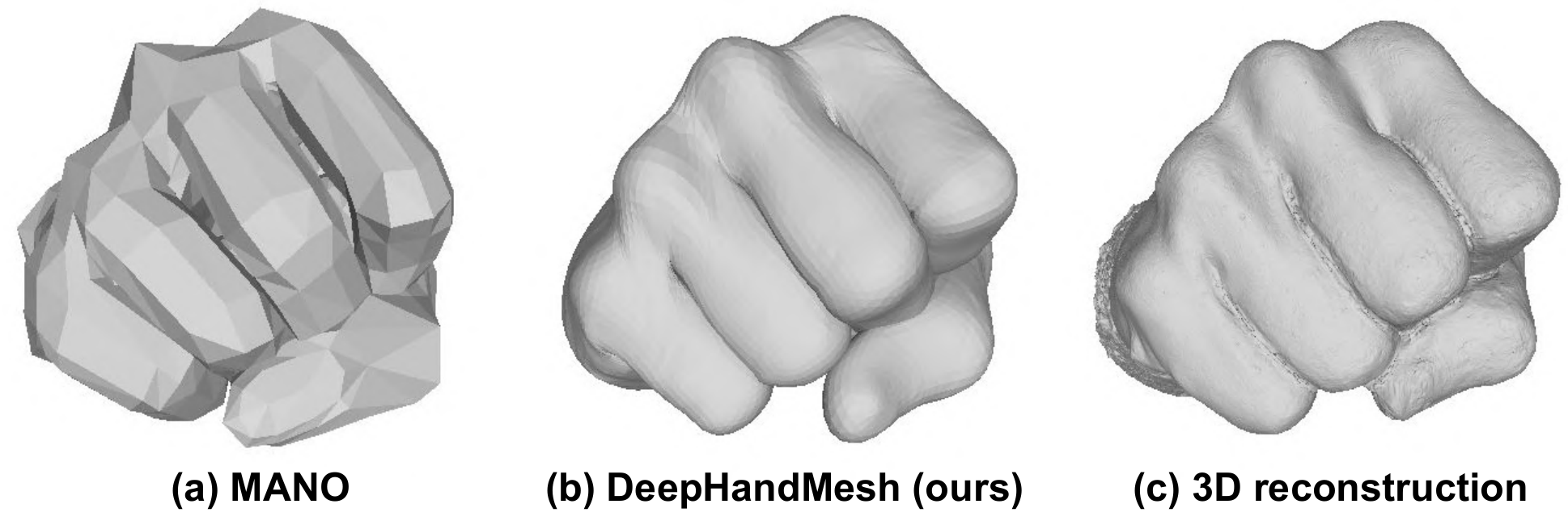}
\end{center}
   \caption{
Qualitative result comparison between (a) MANO~\cite{romero2017embodied}, (b) our DeepHandMesh, and (c) 3D reconstruction~\cite{galliani2015massively}. 
   }
\label{fig:intro_qualitative}
\end{figure}

Although MANO has been widely used for hand pose and geometry estimation~\cite{baek2019pushing,boukhayma20193d,hasson2019learning}, there exist limitations. First, their method requires groundtruth hand meshes  (\textit{i.e.}, the method requires per-vertex supervision to train the linear model). As the hand contains many self-occlusions and self-similarities, existing mesh registration methods~\cite{bogo2014faust,hirshberg2012coregistration} sometimes fail. To obtain the best quality of groundtruth hand meshes, Romero~et al.~\cite{romero2017embodied} manually inspected each registered mesh and discarded failed ones from the training data, which requires extensive manual labor. Second, its fidelity is limited. As MANO uses the hand parts of SMPL~\cite{loper2015smpl}, its resolution is low (\textit{i.e.}, 778 vertices). This low resolution could limit the expressiveness of the reconstructed hand meshes.
Also, MANO consists of linear models, optimized by the classical optimization framework.
As recent deep neural networks (DNNs) that consist of many non-linear modules show noticeable performance in many computer vision and graphics tasks, utilizing the DNNs with recent deep learning optimization techniques can give more robust and stable results. Finally, it does not consider physical interaction between hand parts. A model without consideration of the physical interaction could result in implausible hand deformation, such as penetration between hand parts.

In this paper, we firstly present \textit{DeepHandMesh}, a weakly-supervised deep encoder-decoder framework for high-fidelity hand mesh modeling, that produces high-fidelity hand meshes from single images.
Unlike existing methods such as MANO that require mesh registration for per-vertex supervision (\textit{i.e.}, full supervision), DeepHandMesh utilizes only 3D joint coordinates and multi-view depth maps for supervision (\textit{i.e.}, weak supervision).
Therefore, our method avoids expensive data pre-processing such as registration and manual inspection.
In addition, obtaining the 3D joint coordinates and depth maps is much easier compared with the mesh registration.
The 3D joint coordinates can be obtained from powerful state-of-the-art multi-view 3D human pose estimation methods~\cite{li2019rethinking}, and the depth maps can be rendered from 3D reconstruction~\cite{galliani2015massively} based on the solid mathematical theory about epipolar geometry.
Furthermore, these are independent of topology of a hand model, allowing us to use hand meshes with various topology and to be free from preparing topology-specific data such as registered meshes for each topology.
To achieve high-fidelity hand meshes, DeepHandMesh is based on a DNN and optimized with recent deep learning optimization techniques, which provides more robust and stable results. We also use a high-resolution hand model to benefit from the expressiveness of the DNN. Our DeepHandMesh can replicate realistic hand meshes with details such as creases and skin bulging, as well as holistic hand poses. 
In addition, our newly designed penetration avoidance loss further improves results by enabling our system to replicate physical interaction between hand parts. 
Figure~\ref{fig:intro_qualitative} shows that the proposed DeepHandMesh provides significantly more realistic hand meshes than the existing fully-supervised hand model (\textit{i.e.}, MANO~\cite{romero2017embodied}).

\begin{figure}[t]
\begin{center}
\includegraphics[width=0.6\linewidth]{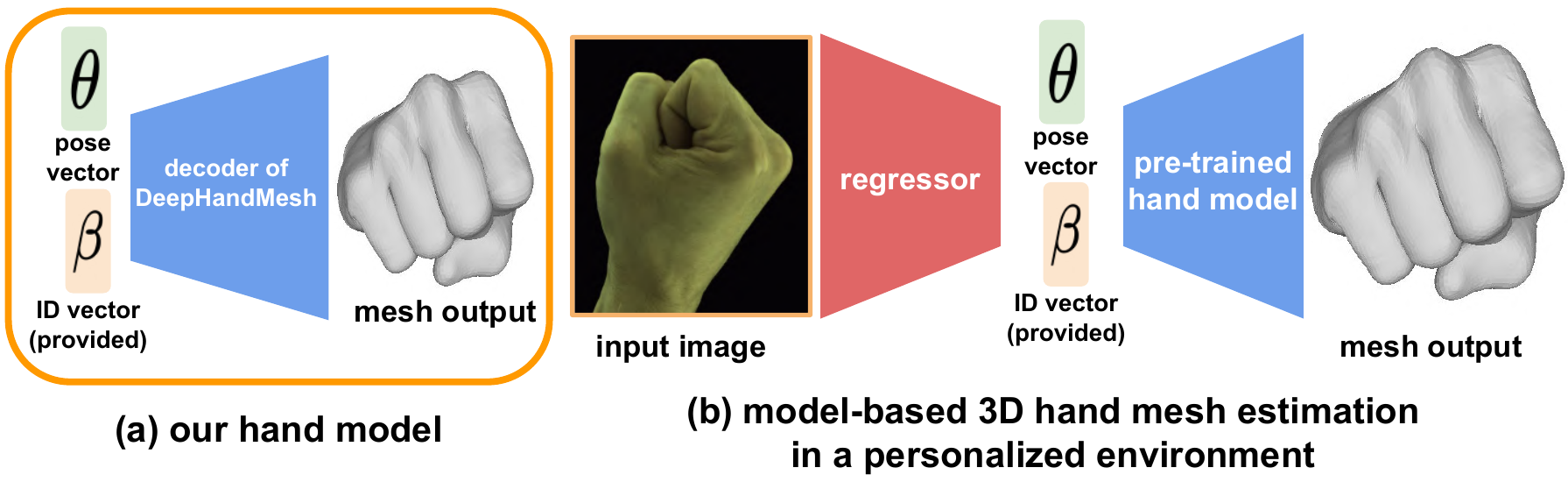}
\end{center}
   \caption{(a) The hand model outputs meshes from the hand model parameters. 
   Our main goal is to train a high-fidelity hand model in a weakly-supervised way.
   (b) The model-based 3D hand mesh estimation system outputs inputs of the hand model and use a pre-trained hand model to produce final hand meshes.}
\label{fig:model_based_methods}
\end{figure}

As learning a high-fidelity hand model only via weak supervisions is a challenging problem, 
we assume a personalized environment (\textit{i.e.}, assume the same subject in the training and testing stage).
We discuss the limitations of the assumption and future research directions in the later section.
To demonstrate the effectiveness of DeepHandMesh for practical purposes, we combine our DeepHandMesh with 3D pose estimation to build a model-based 3D hand mesh estimation system from a single image, as shown in Figure~\ref{fig:model_based_methods}, and train it on a public dataset captured from general environments.
The experimental results show that our DeepHandMesh can be applied to 3D high-fidelity hand mesh estimation from general images in real-time (\textit{i.e.}, 50 fps).

Our contributions can be summarized as follows.
\begin{itemize}
\item We firstly propose a deep learning-based weakly-supervised encoder-decoder framework (DeepHandMesh) that is trained in an end-to-end, weakly-supervised manner for high-fidelity hand mesh modeling.
Our proposed DeepHandMesh does not require labor-intensive manual intervention, such as mesh registration.
\item Our weakly-supervised DeepHandMesh provides significantly more realistic hand meshes than previous fully-supervised hand models. In addition, we newly introduce a penetration avoidance loss, which can make DeepHandMesh firstly reproduce physical interaction between hand parts.
\item We show that our framework can be applied to practical purposes, such as 3D hand mesh estimation from general images in real-time.
\end{itemize}

\section{Related works}

\noindent\textbf{3D hand pose estimation.}
3D hand pose estimation methods can be categorized into depth map-based and RGB-based ones according to their input. Early depth map-based methods are mainly based on a generative approach, which fits a pre-defined hand model to the input depth map by minimizing hand-crafted cost functions~\cite{sharp2015accurate, tang2015opening} using particle swarm optimization~\cite{sharp2015accurate}, iterative closest point~\cite{tagliasacchi2015robust}, or their combination~\cite{qian2014realtime}. 
Most of recent depth map-based methods are based on a discriminative approach, which directly localizes hand joints from an input depth map. Tompson~et al.~\cite{tompson2014real} utilized a neural network to localize hand joints by estimating 2D heatmaps for each hand joint. Ge~et al.~\cite{ge2016robust} extended this method by estimating multi-view 2D heatmaps. Moon~et al.~\cite{moon2018v2v} designed a 3D CNN that takes a voxel representation of a hand as input and outputs a 3D heatmap for each joint. Wan~et al.~\cite{wan2019self} proposed a self-supervised system, which can be trained from only an input depth map. 

The powerful performance of the recent CNN makes 3D hand pose estimation methods work well on RGB images. Zimmermann~et al.~\cite{zimmermann2017learning} proposed a DNN that learns an implicit 3D articulation prior. Mueller~et al.~\cite{mueller2018ganerated} used an image-to-image translation model to generate synthetic hand images for more effective training of a pose prediction model. Cai~et al.~\cite{cai2018weakly} and Iqbal~et al.~\cite{iqbal2018hand} implicitly reconstruct depth map from an input RGB image and estimate 3D hand joint coordinates from it. Spurr~et al.~\cite{spurr2018cross} and Yang~et al.~\cite{yang2019disentangling} proposed variational auto-encoders (VAEs) that learn a latent space of a hand skeleton and appearance.

\noindent\textbf{3D hand shape estimation.}
Panteleris~et al.~\cite{panteleris2018using} fitted a pre-defined hand model by minimizing reprojection errors of 2D joint locations w.r.t. hand landmarks detected by OpenPose~\cite{simon2017hand}. Ge~et al.~\cite{ge20193d} proposed a graph convolution-based network which directly estimates vertices of a hand mesh. Many recent methods are based on the MANO hand model. They train their new encoders and use a pre-trained MANO model as a decoder to generate hand meshes. Baek~et al.~\cite{baek2019pushing} trained their network to estimate input vectors of the MANO model using neural renderer~\cite{kato2018neural}. Boukhayma~et al.~\cite{boukhayma20193d} proposed a network that takes a single RGB image and estimates pose and shape vectors of MANO. Their network is trained by minimizing the distance of the estimated hand joint locations and groundtruth. Recently, Zimmermann~et al.~\cite{zimmermann2019freihand} proposed a marker-less captured 3D hand pose and mesh dataset.

\noindent\textbf{3D hand model.}
MANO~\cite{romero2017embodied} is the most widely used hand model. It takes pose and shape vectors (\textit{i.e.}, relative rotation of hand joint w.r.t. its parent joint and principal component analysis coefficients of hand shape space, respectively) as inputs and outputs deformed mesh using LBS and per-vertex correctives. It is trained from registered hand meshes in a fully-supervised way by minimizing the per-vertex distance between the output and the groundtruth hand meshes. Recently, Kulon~et al.~\cite{kulon2019single} proposed a hand model that takes a mesh latent code and outputs a hand mesh using mesh convolution. To obtain the groundtruth meshes, they registered their new high-resolution hand model to 3D joint coordinates of Panoptic dome dataset~\cite{simon2017hand}. They also compute a distribution of valid poses from the hand meshes registered to $\sim$1000 scans from the MANO dataset. They use this distribution to sample groundtruth hand meshes and train their hand model in a fully-supervised way using per-vertex mesh supervision.

All the above 3D hand models rely on mesh supervision (\textit{i.e.}, trained by minimizing the per-vertex distance between output and groundtruth hand mesh) during training. In contrast, our DeepHandMesh is trained in a weakly-supervised setting, which does not require any groundtruth hand meshes. Although ours is trained without mesh supervision, it successfully reconstructs significantly more high-fidelity hand meshes, including creases and skin bulging, compared with previous hand models. Also, our DeepHandMesh is the first hand model that can replicate physical interaction between hand parts. This is a significant advancement compared with previous hand models.

\begin{figure}[t]
\begin{center}
\includegraphics[width=0.8\linewidth]{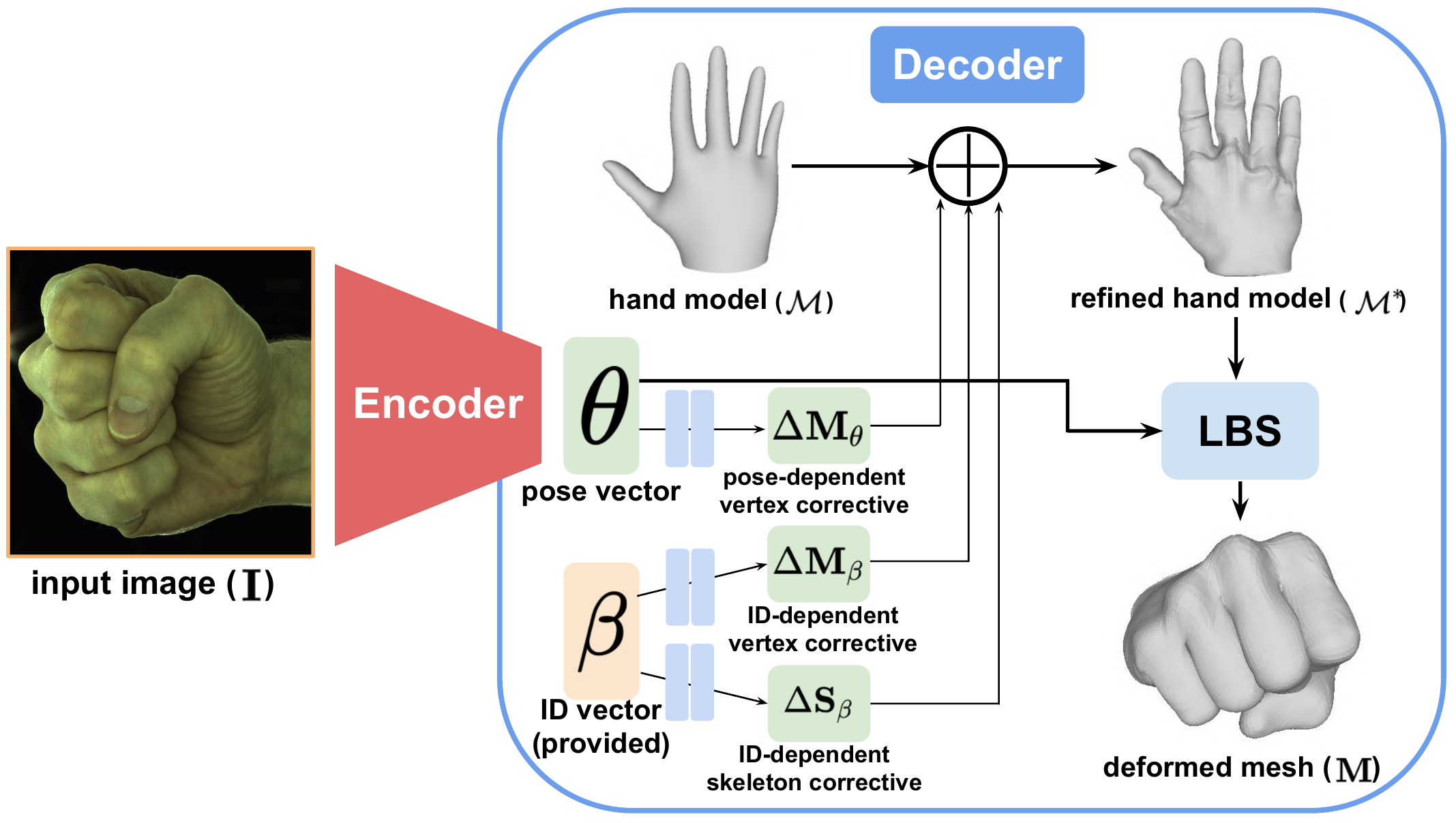}
\end{center}
   \caption{Overall pipeline of the proposed DeepHandMesh.}
\label{fig:overall_pipeline}
\end{figure}

\section{Hand model}
Our hand model is defined as $\mathcal{M}=\{ \mathbf{\bar M}, \mathbf{S} ; \mathbf{W}, \mathcal{H} \}$. $\mathbf{\bar M} = [\mathbf{\bar m}_1, \ldots, \mathbf{\bar m}_V]^T \in \mathbb{R}^{V \times 3}$ denotes vertex coordinates of a zero-pose template hand mesh, where $\mathbf{\bar m}_v$ is 3D coordinates of $v$th vertex of $\mathbf{\bar M}$. 
$V$ denotes the number of vertices.
$\mathbf{S} \in \mathbb{R}^{J \times 3}$ means the translation vector of each hand joint from its parent joint, where $J$ is the number of joints. $\mathbf{W} \in \mathbb{R}^{V \times J}$ denotes skinning weights for LBS. Finally, $\mathcal{H}$ denotes a hand joint hierarchy. Our template hand model is prepared by artists. 
The parameters on the right of the semicolon do not change during training. 
Thus, we omit them hereafter for simplicity.

\section{Encoder}
\label{sec:Encoder}

\subsection{Hand pose vector}
\label{ssec:HandPoseVector}
The encoder takes a single RGB image of a hand $\mathbf{I}$ and estimates its hand pose vector $\theta \in \mathbb{R}^{N_\text P}$, where $N_\text P=28$ denotes the degrees of freedom (DOFs) of it.
Among all the DOFs of the hand joint rotation $3J$, we selected $N_\text P$ DOFs based on the prior knowledge of human hand anatomical property and the hand models of ~\cite{zhou2016model,yuan2017bighand2}. For the enabled DOFs, the estimated hand pose vector is used as a relative Euler angle w.r.t. its parent joint. We set all the disabled DOFs to zero and fixed them during the optimization.

\subsection{Network architecture}
\label{ssec:NetworkArchitecture}
Our encoder consists of ResNet-50~\cite{he2016deep} and two fully-connected layers.
The ResNet extracts a hand image feature from the input RGB image $\mathbf{I}$. 
Then, the extracted feature is passed to the two fully-connected layers, which outputs the hand pose vector $\theta$.
The hidden activation size of the fully-connected layers is 512, and the ReLU activation function is used after the first fully-connected layer.
To ensure $\theta$ in the range of (-$\pi$, $\pi$), we apply a hyperbolic tangent activation function at the output of the second fully-connected layer and multiply it by $\pi$.

\section{Decoder}
\label{sec:Decoder}

\subsection{Hand model refinement}
\label{ssec:HandModelRefinement}
To replicate details on the hand model,
we designed the decoder to estimate three correctives from a pre-defined identity vector $\beta \in \mathbb{R}^{N_\text I}$ and an estimated hand pose vector $\theta$, inspired by~\cite{loper2015smpl,romero2017embodied}, as shown in Figure~\ref{fig:overall_pipeline}. 
As the proposed DeepHandMesh assumes a personalized environment (\textit{i.e.}, assumes the same subject in the training and testing stage), we pre-define $\beta$ as a $N_\text I=32$ dimensional randomly initialized normal Gaussian vector for each subject.
$\beta$ is fixed during training and testing.
Note that DeepHandMesh does not require a personalized hand model to be given. Rather, it personalizes an initial hand mesh for a training subject during training.

\begin{figure}[t]
\begin{center}
\includegraphics[width=0.8\linewidth]{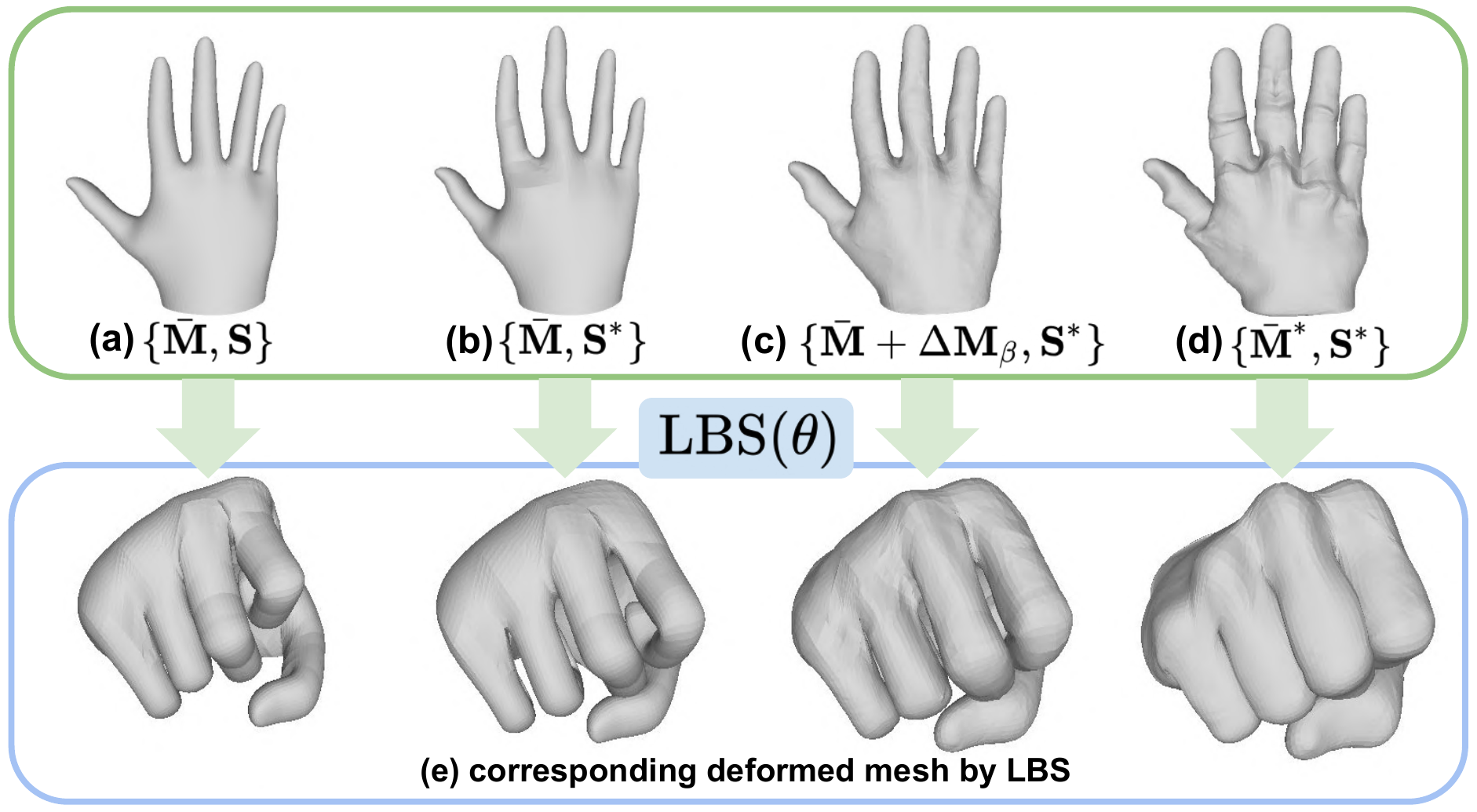}
\end{center}
   \caption{(a)-(d): Visualized hand model refined by different combinations of correctives. (e): Deformed hand model using LBS.}
\label{fig:each_component}
\end{figure}

The first corrective is identity-dependent skeleton corrective $\Delta \mathbf{S}_{\beta} \in \mathbb{R}^{J \times 3}$. As hand shape and size vary for each person, 3D joint locations can be different for each person. To personalize $\mathbf{S}$ to a training subject, we build two fully-connected layers in our decoder and estimate $\Delta \mathbf{S}_{\beta}$ from the pre-defined identity code $\beta$. The hidden activation size of the fully-connected layer is 256. The estimated $\Delta \mathbf{S}_{\beta}$ is added to $\mathbf{S}$, yielding $\mathbf{S}^*$. Figure~\ref{fig:each_component} (b) shows the effect of $\Delta \mathbf{S}_{\beta}$.

The second corrective is identity-dependent per-vertex corrective $\Delta \mathbf{M}_{\beta} \in \mathbb{R}^{V \times 3}$. In addition to the 3D joint locations, hand shape such as finger thickness is also different for each person. To cope with the shape difference, we build two fully-connected layers and estimate $\Delta \mathbf{M}_{\beta}$ from the identity code $\beta$. The hidden activation size of the fully-connected layer is 256. The estimated $\Delta \mathbf{M}_{\beta}$ is added to $\bar{\mathbf{M}}$. Figure~\ref{fig:each_component} (c) shows the effect of $\Delta \mathbf{M}_{\beta}$.

The last corrective is pose-dependent per-vertex corrective $\Delta \mathbf{M}_{\theta} \in \mathbb{R}^{V \times 3}$. 
When making a pose (\textit{i.e.}, $\theta$ varies), local deformation of hand geometry such as skin bulging and crease appearing/disappearing also occurs.
To recover such phenomena, we build two fully-connected layers to estimate $\Delta \mathbf{M}_{\theta}$ from the hand pose vector $\theta$. The hidden activation size of the fully-connected layer is 256. The estimated $\Delta \mathbf{M}_{\theta}$ is added to $\bar{\mathbf{M}}$. For stable training, we do not back-propagate gradient from $\Delta \mathbf{M}_{\theta}$ through $\theta$. Figure~\ref{fig:each_component} (d) shows the effect of $\Delta \mathbf{M}_{\theta}$.

The final refined hand model $\mathcal{M}^*$ is obtained as follows:
\begin{gather*}
\bar{\mathbf{M}}^* = \bar{\mathbf{M}} + \Delta \mathbf{M}_{\theta} + \Delta \mathbf{M}_{\beta}, \, \mathbf{S}^* = \mathbf{S} + \Delta \mathbf{S}_{\beta}, \\
\mathcal{M}^* = \{\bar{\mathbf{M}}^*, \mathbf{S}^* \}.
\end{gather*}

\subsection{Hand model deformation}
\label{ssec:HandModelDeformation}
We first perform 3D rigid alignment from the hand model space to the dataset space for the global alignment using the wrist and finger root positions. Then, we use the LBS algorithm to holistically deform our hand model. LBS is a widely used algorithm to deform a mesh according to linear combinations of joint rigid transformation~\cite{loper2015smpl,romero2017embodied}. Specifically, each vertex $\mathbf{m}_v$ of a deformed hand mesh $\mathbf{M} \in \mathbb{R}^{V \times 3}$ is obtained as follows:
\begin{equation}
\begin{split}
\mathbf{m}_v = & (\mathbf{I}_3,\mathbf{0}) \cdot \sum_{j=1}^{J}w_{v,j}\mathbf{T}_j(\theta,\mathbf{S}^*;\mathcal{H}) \begin{pmatrix} \bar{\mathbf{m}}_v^* \\ 1 \end{pmatrix} \\
= & \mathrm{LBS}(\theta,\mathbf{S}^*,\bar{\mathbf{m}}_v^*), v = 1,\ldots,V,
\end{split}
\end{equation}
where $\mathbf{T}_j(\theta,\mathbf{S}^*;\mathcal{H}) \in \mathrm{SE(3)}$ denotes transformation matrix for joint $j$. It encodes the rotation and translation from the zero pose to the target pose, constructed by traversing the hierarchy $\mathcal{H}$ from the root to $j$. $w_{v,j}$ and $\bar{\mathbf{m}}_v^*$ denote $j$th joint of $v$th vertex skinning weight from $\mathbf{W}$ and $v$th vertex coordinate from $\bar{\mathbf{M}}^*$, respectively. The visualization of a deformed mesh is shown in Figure~\ref{fig:each_component} (e).

\section{Training DeepHandMesh}
\label{sec:loss}

We use four loss functions to train DeepHandMesh. The \textbf{Pose loss} and \textbf{Depth map loss} are responsible for the weak supervision. The \textbf{Penetration loss} helps to reproduce physical interaction between hand parts and the \textbf{Laplacian loss} acts as a regularizer to make output hand meshes smooth. 

\noindent\textbf{Pose loss.} We perform forward kinematics from the estimated hand pose vector $\theta$ and refined skeleton $\mathbf{S}^*$ to get the 3D coordinates of the hand joints $\mathbf{P} = [\mathbf{p}_1, \ldots, \mathbf{p}_J]^T \in \mathbb{R}^{J \times 3}$. We minimize $L1$ distance between the estimated and the groundtruth coordinates. The pose loss is defined as follows:
\noindent$L_\text{pose} = \frac{1}{J} \sum_{j=1}^{J} || \mathbf{p}_j - \mathbf{p}_j^* ||_1$,
where $*$ indicates the groundtruth.

\noindent\textbf{Depth map loss.} We render 2D depth maps $\mathcal{D} = (\mathbf{D}_1, \ldots, \mathbf{D}_{C_\text{out}})$ of $\mathbf{M}$ from randomly selected $C_\text{out}$ target views, and minimize $\mathrm{Smooth}_{L1}$ distance~\cite{girshick2015fast} between the rendered and the groundtruth depth maps following Ge~et al.~\cite{ge20193d}. To make the depth map loss differentiable, we use Neural Renderer~\cite{kato2018neural}. The depth map loss is defined as follows:
\noindent$L_\text{depth} = \frac{1}{C_\text{out}} \sum_{c=1}^{C_\text{out}} \delta_c \left(\mathrm{Smooth}_{L1} (\mathbf{D}_c, \mathbf{D_c^*})\right)$,
where $*$ indicates the groundtruth. $\delta_c$ is a binary map whose pixel value of each grid is one if it is foreground (\textit{i.e.}, a depth value is defined in $\mathbf{D}_c$ and $\mathbf{D}_c^*$), and zero otherwise.

\begin{figure}[t]
\begin{center}
   \includegraphics[width=0.5\linewidth]{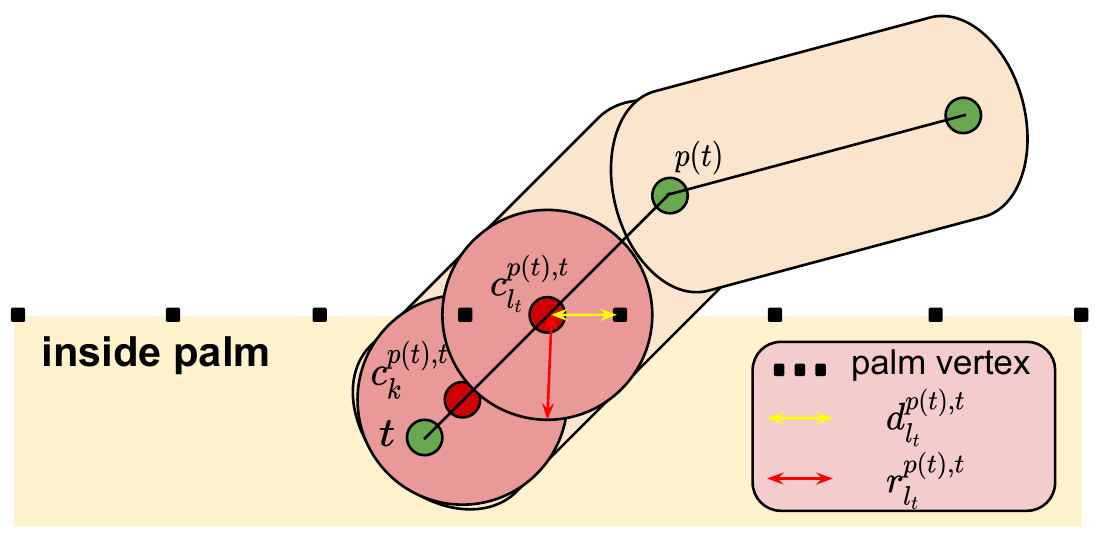}
\end{center}
   \caption{Visualized example of penetration between a finger and palm.}
\label{fig:nr_penet_example}
\end{figure}

\noindent\textbf{Penetration loss.} To penalize penetration between hand parts, we introduce two penetration avoidance regularizers. We consider the fingers as rigid hand parts and the palm as a non-rigid hand part. The regularizers are designed for each of the rigid and non-rigid parts.  

For the rigid parts (\textit{i.e.}, fingers), we use a regularizer similar to that in Wan~et al.~\cite{wan2019self}, which represents each rigid part with a combination of spheres. Specifically, we compute a pair of the center and radius of spheres $\{s_k^{p(j),j} = (c_k^{p(j),j},r_k^{p(j),j})\}_{k=1}^K$ between joint $j$ and its parent joint $p(j)$, where $K =10$ denotes the number of spheres between the adjacent joints. The center $c_k^{p(j),j}$ is computed by linearly interpolating $\bar{\mathbf{p}}_{p(j)}$ and $\bar{\mathbf{p}}_j$, where $c_1^{p(j),j} = \bar{\mathbf{p}}_{p(j)}$ and $c_K^{p(j),j} = \bar{\mathbf{p}}_j$. $\bar{\mathbf{p}}_j$ denotes the 3D coordinate of hand joint $j$ obtained from forward kinematics using $\theta = \vec{0}$ and $\mathbf{S}^*$. Each radius $r_k^{p(j),j}$ is obtained by calculating the distance between $c_k^{p(j),j}$ and the closest vertex in $\mathrm{LBS}(\vec{0},\mathbf{S}^*,\bar{\mathbf{M}}^*)$. Given these spheres, the penetration avoidance term between the rigid hand parts $L_\text{penet}^\text{r}$ is defined as follows:
\begin{equation}
L_\text{penet}^\text{r} = \sum_{\mathclap{\substack{k,k\prime \\
                        j \neq j\prime, p(j\prime)\\
                        j\prime \neq p(j)}}} \max(r_k^{p(j),j}+r_{k\prime}^{p(j\prime),j\prime} - ||c_k^{p(j),j} - c_{k\prime}^{p(j\prime),j\prime}||_2, 0),
\end{equation}
which indicates that the distances of any pairs of the spheres except the ones associated with adjacent joints are enforced to be greater than the sum of the radii of the paired spheres.
This prevents overlap between the spheres, thus avoiding penetration between the rigid parts.

However, $L_\text{penet}^\text r$ does not help prevent penetration at the non-rigid hand part (\textit{i.e.}, the palm).
The underlying assumption of $L_\text{penet}^\text r$ is that surface geometry can be approximated by many spheres. While this assumption holds for the fingers due to the cylindrical shape, it does not often hold for the palm, \textit{i.e.}, the spheres along the joints in the palm cannot approximate the palm surface particularly when pose-dependent corrective replicating skin bulging is applied.
Additionally, $L_\text{penet}^\text r$ does not produce surface deformation, \textit{e.g.}, finger-palm collision often makes large deformation to the palm surface.
$L_\text{penet}^\text r$ does not help replicate such deformation.

To address those limitations, we propose a new penetration avoidance term $L_\text{penet}^\text{nr}$ for the non-rigid hand part. For this, we only consider penetration between fingertips and palm as illustrated in Figure~\ref{fig:nr_penet_example}. Among $\mathbf{M}$, vertices whose most dominant joint in the skinning weight $\mathbf{W}$ is the palm are considered as ones for the palm $\mathbf{M}_\gamma$. Then, the distance between $c^{p(t),t}_k$ and $\mathbf{M}_\gamma$ is calculated, where $t$ is one of fingertip joints. Among the distances, the shortest one is denoted as $d^{p(t),t}_k$. If there exists $l_t$ where $d^{p(t),t}_{l_t}$ is smaller than $r^{p(t),t}_{l_t}$, we consider that $c^{p(t),t}_{l_t}$ penetrates $\mathbf{M}_\gamma$. If there are more than one $l_t$, we use the one closest to the $p(t)$, which is considered as a starting point of penetration. Based on human hand anatomical property, we can conclude that the spheres from $l_t$ to the fingertip $\{s^{p(t),t}_k\}_{k={l_t}}^K$ are penetrating $\mathbf{M}_\gamma$. 
Then, we enforce $\{d^{p(t),t}_k\}_{k={l_t}}^K$ to be the same as $\{r^{p(t),t}_k\}_{k={l_t}}^K$. The penetration avoidance term for the non-rigid hand part $L_\text{penet}^\text{nr}$ is defined as follows:
\begin{equation}
L_\text{penet}^\text{nr} = \sum_{t} g(t),
\end{equation}
\begin{equation}
\text{where } g(t) = 
\begin{cases}
\sum_{k=l_t}^K |d^{p(t),t}_k - r^{p(t),t}_k|,& \text{if } l_t \text{~exists} \\
    0,& \text{otherwise}.
\end{cases}
\end{equation}

The final penetration avoidance loss function is defined as follows:
\noindent$L_\text{penet} = L_\text{penet}^\text r + \lambda_\text{nr} L_\text{penet}^\text{nr}$,
where $\lambda_\text{nr} = 5$.

\noindent\textbf{Laplacian loss.}
To preserve local geometric structure of the deformed mesh based on the mesh topology,
we add a Laplacian regualarizer~\cite{liu2019soft} as follows:
\noindent$L_\text{lap} = \frac{1}{V} \sum_{v=1}^V \left(\mathbf{m}_v - \frac{1}{||\mathcal{N}(v)||} \sum_{v^\prime \in \mathcal{N}(v)} \mathbf{m}_{v^\prime} \right)$,
where $\mathcal{N}(v)$ denotes neighbor vertices of $\mathbf{m}_v$.

Our DeepHandMesh is trained in an end-to-end manner. Note that although our DeepHandMesh is trained without per-vertex mesh supervision, it can be trained with a single regularizer $L_\text{lap}$. The total loss function $L$ is defined as follows:
\noindent$L = L_\text{pose} + L_\text{depth} + L_\text{penet} + \lambda_\text{lap} L_\text{lap}$,
where $\lambda_\text{lap}=5$. 

\section{Implementation details}

PyTorch~\cite{paszke2017automatic} is used for implementation. The ResNet in the encoder is initialized with the publicly released weights pre-trained on the ImageNet dataset~\cite{russakovsky2015imagenet}, and the weights of the remaining part are initialized by Gaussian distribution with zero mean and $\sigma=0.01$. The weights are updated by the Adam optimizer~\cite{kingma2014adam} with a mini-batch size of 32. The number of rendering views is $C_\text{out}$ = 6.
We use 256$\times$256 as the size of $\mathbf{I}$ and depth maps of $\mathcal{D}$. We observed that changing $C_\text{out}$ and resolution of $\mathbf{I}$ and depth maps of $\mathcal{D}$ does not affect much the quality of the resulting mesh. The number of vertices in our hand model is 12,553. We train our DeepHandMesh for 35 epochs with a learning rate of $10^{-4}$. The learning rate is reduced by a factor of 10 at the \nth{30} and \nth{32} epochs. We used four NVIDIA Titan V GPUs for training, which took 9 hours.
Both the encoder and decoder of our DeepHandMesh run at 100 fps, yielding real-time performance (50 fps).

\section{Experiment}

\subsection{Dataset} \label{sec:dataset}
We used the same data capture studio with Moon~et al.~\cite{moon2020inter}.
The experimental image data was captured by 80 calibrated cameras capable of synchronously capturing images with 4096 $\times$ 2668 pixels at 30 frames per second. All cameras lie on the front, side, and top hemisphere of the hand and are placed at a distance of about one meter from it. 
During capture, each subject was instructed to make a pre-defined set of 40 hand motions and 15 conversational gestures. 
We pre-processed the raw video data by performing multi-view 3D hand pose estimation~\cite{li2019rethinking} and multi-view 3D reconstruction~\cite{galliani2015massively}. 
We split our dataset into training and testing sets.
The training set contains 404K images per subject with the 40 pre-defined hand poses, and the test set contains 80K images per subject with the 15 conversational gestures.
There are four subjects (one female and three males), and we show more detailed description and various examples of our dataset in the supplementary material.

\begin{figure}[t]
\begin{center}
   \includegraphics[width=1.0\linewidth]{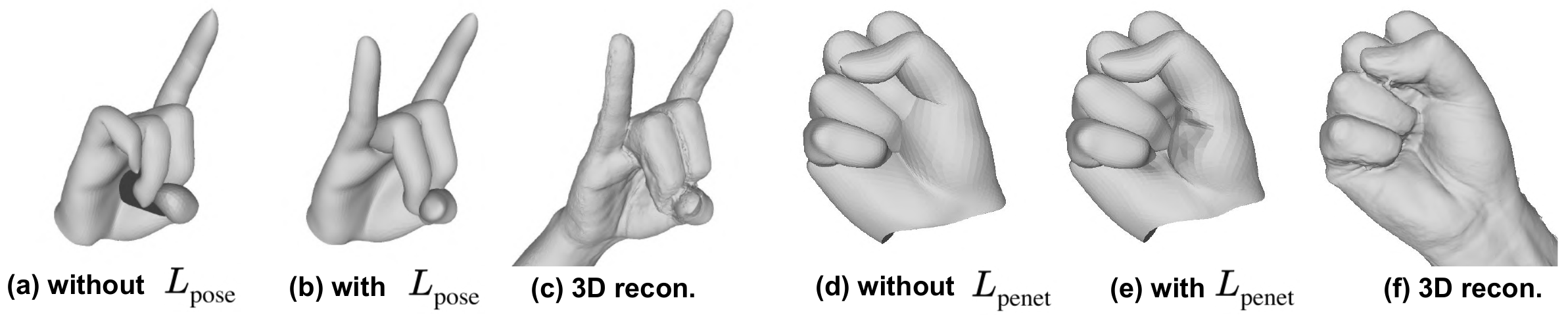}
\end{center}
   \caption{(a)-(c): Deformed hand mesh trained without and with $L_\text{pose}$, and corresponding 3D reconstruction~\cite{galliani2015massively}. (d)-(f): Deformed hand mesh trained without and with $L_\text{penet}$, and corresponding 3D reconstruction~\cite{galliani2015massively}.}
\label{fig:effect_of_loss}
\end{figure}

\begin{figure}[t]
\begin{center}
   \includegraphics[width=1.0\linewidth]{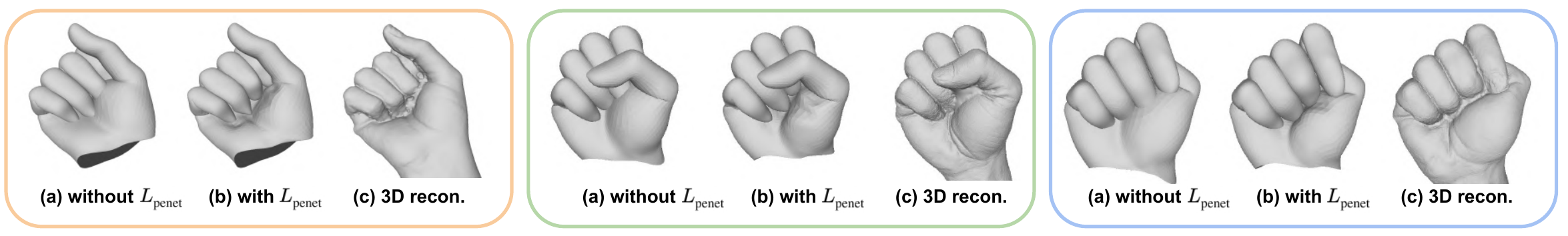}
\end{center}
   \caption{Deformed hand mesh trained without and with $L_\text{penet}$, and corresponding 3D reconstruction~\cite{galliani2015massively}.}
\label{fig:effect_of_penet}
\end{figure}

\begin{figure}[t]
\begin{center}
   \includegraphics[width=0.7\linewidth]{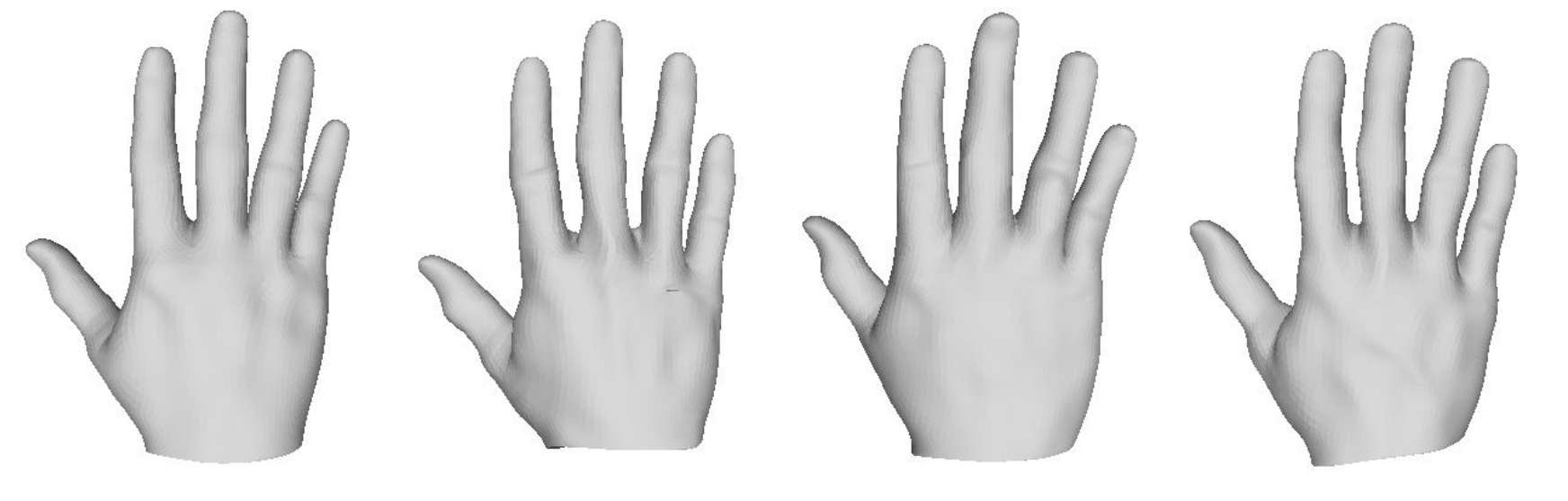}
\end{center}
   \caption{Visualized hand models of zero pose from different subjects.}
\label{fig:effect_of_id_corrective}
\end{figure}

\begin{figure}[t]
\begin{center}
   \includegraphics[width=1.0\linewidth]{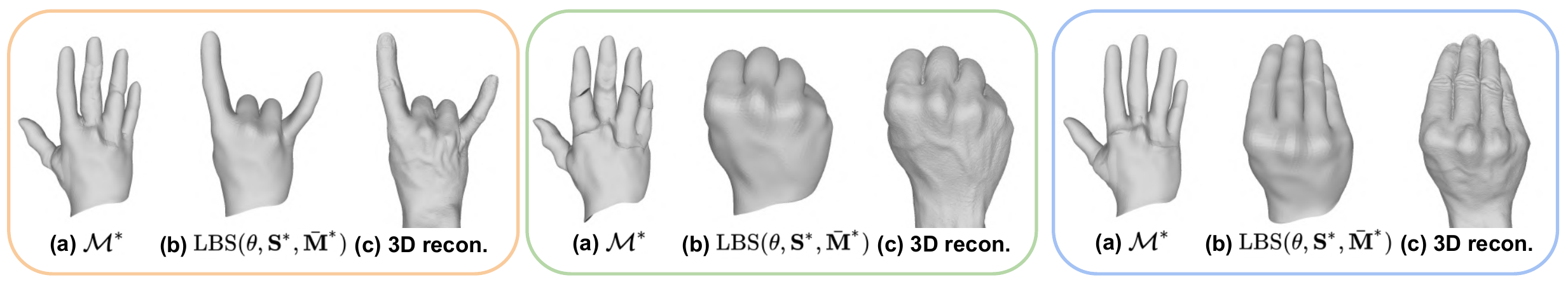}
\end{center}
   \caption{(a) Refined hand model, (b) deformed hand mesh, and (c) 3D reconstruction~\cite{galliani2015massively} from various hand poses of a one subject.}
\label{fig:effect_of_pose_corrective}
\end{figure}

\subsection{Ablation study}

\noindent\textbf{Effect of each loss function.}
To investigate the effect of each loss function, we visualize test results from models trained with different combinations of loss functions in Figure~\ref{fig:effect_of_loss}. In the figure, (a), (b), (d), and (e) are the results of our DeepHandMesh, and (c) and (f) are the results of 3D reconstruction~\cite{galliani2015massively}, respectively.

The model trained without $L_\text{pose}$ (a) gives wrong joint locations. Also, there are severe artifacts at occluded hand regions (\textit{e.g.}, the black area on the palm region) because of skin penetration. This is because $L_\text{depth}$ cannot back-propagate gradients through occluded areas. In contrast, $L_\text{pose}$ can give gradients at the invisible regions, which makes more stable and accurate results, as shown in (b). 
The model trained without $L_\text{penet}$ (d) cannot prevent penetration between fingers and palm. However, $L_\text{penet}$ penalizes this, and the fingertip locations are placed more plausibly, and the palm vertices are deformed according to the physical interaction between the fingers and palm, as shown in (e).
Figure~\ref{fig:effect_of_penet} additionally shows the effectiveness of the proposed $L_\text{penet}$.

\begin{figure}[t]
\begin{center}
\includegraphics[width=0.8\linewidth]{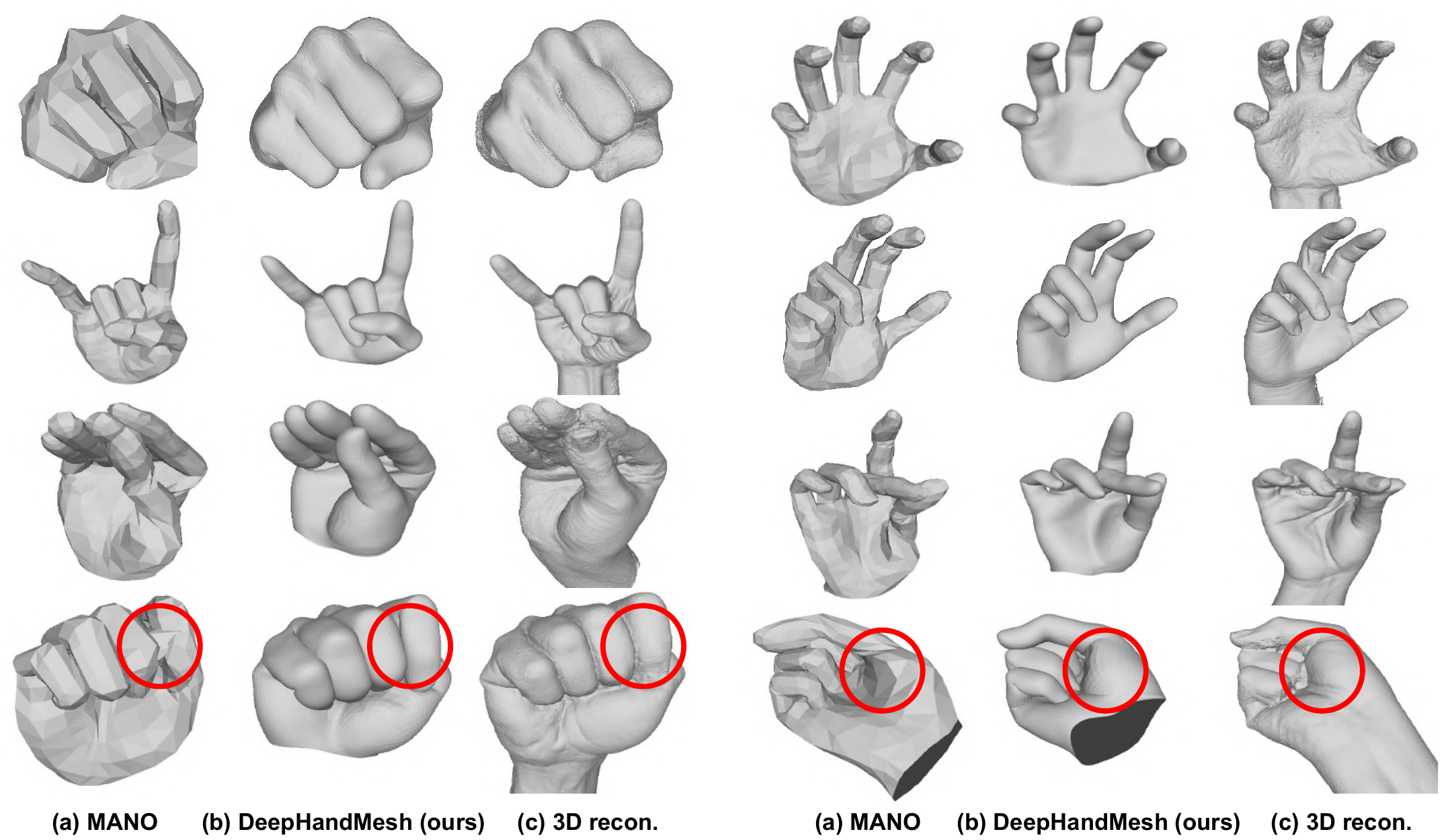}
\end{center}
   \caption{Estimated hand mesh comparison from various hand poses and subjects with the state-of-the-art method. The red circles in the last row show physical interaction between hand parts.}
\label{fig:comparison_mano}
\end{figure}

\noindent\textbf{Effect of identity-dependent correctives.}
To demonstrate the effectiveness of our identity-dependent corrective (\textit{i.e.}, $\Delta \mathbf{S}_\beta$ and $\Delta \mathbf{M}_\beta$), we visualize how our DeepHandMesh handles different identities in Figure~\ref{fig:effect_of_id_corrective}. The figures are drawn by setting $\theta = \vec{0}$ to normalize hand pose. As the figures show, our identity-dependent corrective successfully personalizes the initial hand model to each subject by adjusting the hand bone lengths and skin.

\noindent\textbf{Effect of pose-dependent corrective.}
To demonstrate the effectiveness of our pose-dependent per-vertex corrective $\Delta \mathbf{M}_\theta$, we visualize the hand meshes of different poses in Figure~\ref{fig:effect_of_pose_corrective}. All the hand meshes are from the same subject to normalize identity. For each hand pose, (a) shows the hand model after model refinement with zero pose. (b) shows deformed (a) using LBS, and (c) shows 3D reconstruction meshes. As the figure shows, our pose-dependent correctives successfully recover details according to the poses. Note that in (b), we approximately reproduced local deformation based on the blood vessels.

\subsection{Comparison with state-of-the-art methods}
We compare our DeepHandMesh with widely used hand model MANO~\cite{romero2017embodied} on our dataset. For comparison, we train a model whose encoder is the same one as ours, and decoder is the pre-trained MANO model.
The pre-trained MANO model is fixed during the training, and we use the same loss functions as ours. 
We pre-defined identity code $\beta$ for each subject and estimate the shape vector of MANO from the code using two fully-connected layers to compare both models in the personalized environment. Figure~\ref{fig:comparison_mano} shows the proposed DeepHandMesh provides significantly more realistic hand mesh from various hand poses and identities. In the last row, MANO suffers from the unrealistic physical interaction between hand parts such as finger penetration and flat palm skin. In contrast, our DeepHandMesh does not suffer from finger penetration and can replicate physical interaction between finger and palm skin. Table~\ref{table:comparison_mano} shows the 3D joint coordinate distance error and mesh vertex error from the closest point on the 3D reconstruction meshes for unseen hand poses, indicating that our DeepHandMesh outperforms MANO on the unseen hand pose images. 
For more comparisons, we experimented with lower-resolution hand mesh in the supplementary material.

We found that comparisons between DeepHandMesh and MANO with publicly available 3D hand datasets~\cite{zhang20163d,zimmermann2017learning} were difficult because DeepHandMesh assumes a personalized environment (\textit{i.e.}, assumes the same subject in training and testing stages). 
However, we believe the qualitative and quantitative comparisons in Figure~\ref{fig:comparison_mano} and Table~\ref{table:comparison_mano} still show the superiority of the proposed DeepHandMesh.

\begin{table}[t]
\small
\centering
\setlength\tabcolsep{1.0pt}
\def\arraystretch{1.1}
\scalebox{1.0}{
\begin{tabular}{C{4.5cm}C{2.0cm}C{2.0cm}C{2.0cm}}
\specialrule{.1em}{.05em}{.05em}
methods & $\mathbf{P}_\text{err}$ (mm) & $\mathbf{M}_\text{err}$ (mm) \\ \hline
MANO &  13.81  & 8.93 \\
\textbf{DeepHandMesh (Ours)} &  \textbf{9.86} &  \textbf{6.55} \\ \hline
\end{tabular}
}
\caption{3D joint distance error $\mathbf{P}_\text{err}$ and mesh vertex error $\mathbf{M}_\text{err}$ comparison between MANO and DeepHandMesh on test set consists of unseen hand poses.}
\label{table:comparison_mano}
\end{table}

\subsection{3D hand mesh estimation from general images}
\begin{figure}[t]
\begin{center}
\includegraphics[width=0.9\linewidth]{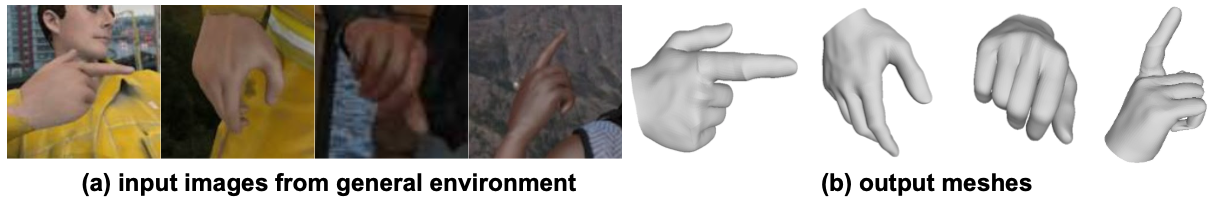}
\end{center}
   \caption{3D hand mesh estimation results from general images.}
\label{fig:general_img}
\end{figure}

To demonstrate a use case of DeepHandMesh for general images, we developed a model-based 3D hand mesh estimation system based on DeepHandMesh.
Figure~\ref{fig:general_img} shows that our model-based 3D hand mesh estimation system generates realistic hand meshes without mesh supervision from the test set of the RHD~\cite{zimmermann2017learning}.
For this, we first pre-trained DeepHandMesh, and replaced its encoder with a randomly initialized one that has exactly the same architecture with our encoder, as illustrated in Figure~\ref{fig:model_based_methods}.
We trained the new encoder on the training set of RHD, by minimizing $L_\text{pose}$.
The RHD dataset contains 44K images synthesized by animating the 3D human models. 
During the training, the decoder is fixed, which is a similar training strategy with that of MANO-based 3D hand mesh estimation methods~\cite{baek2019pushing,boukhayma20193d}.
As our DeepHandMesh assumes a personalized environment, we used a groundtruth bone length to adjust a bone length of the output 3D joint coordinates.
The inputs of the decoder are joint rotations and identity code without any image appearance information like MANO; therefore, the decoder can easily generalize to general images, although it is trained on the data captured from the controlled environment.

\section{Discussion}
Our DeepHandMesh assumes a personalized environment.
Future work should consider cross-identity hand mesh modeling by estimating the Gaussian identity code.
However, training cross-identity hand mesh model in a weakly-supervised way is very hard.
As MANO is trained in a fully-supervised way, they could perform principal component analysis (PCA) on the groundtruth hand meshes in zero-pose and model the identity as coefficients of the principal components.
On the other hand, there is no groundtruth mesh under the weakly-supervised setting, therefore performing PCA on meshes is not possible.
Generative models (\textit{e.g.}, VAE) can be designed to learn a latent space of identities from registered meshes like~\cite{jiang2019disentangled}; however, training a generative model in a weakly supervised way without registered meshes also remains challenging.
We believe the extension of DeepHandMesh to handle cross-identity in a weakly-supervised setting could be an interesting future direction.

\section{Conclusion}
We presented a novel and powerful weakly-supervised deep encoder-decoder framework, DeepHandMesh, for high-fidelity hand mesh modeling. In contrast to the previous hand models~\cite{romero2017embodied,kulon2019single}, DeepHandMesh is trained in a weakly-supervised setting; therefore, it does not require groundtruth hand mesh. Our model successfully generates more realistic hand mesh compared with the previous fully-supervised hand models. The newly introduced penetration avoidance loss makes the result even more realistic by replicating physical interactions between hand parts.

\section*{Acknowledgments} This work was partially supported by the Next-Generation Information Computing Development Program (NRF-2017M3C4A7069369) and the Visual Turing Test project (IITP-2017-0-01780) funded by the Ministry of Science and ICT of Korea.

\clearpage

\begin{center}
\textbf{\large Supplementary Material of \enquote{DeepHandMesh: \\ A Weakly-supervised Deep Encoder-Decoder Framework for High-fidelity Hand Mesh Modeling}}
\end{center}

In this supplementary material, we present more experimental results that could not be included in the main manuscript due to the lack of space.

\section{Texture loss}~\label{sec:texture}
Although the overall shape of the hand mesh is close to the groundtruth, some vertices can move inconsistently across time steps. To prevent this, we employ texture consistency loss, similar to ~\cite{yoon2019self,wei2019vr}. Specifically, we first train the DeepHandMesh and obtain a hand mesh of a neutral pose, which is considered as the easiest pose to estimate. Then, the mesh is used to unwrap neutral pose RGB images from all views of our dataset to a 1024$\times$1024 UV texture $\mathbf{T}^*$ using Poisson reconstruction~\cite{kazhdan2006poisson}. We use $\mathbf{T}^*$ as a groundtruth texture and force texture $\mathbf{T}$ of each iteration to be the same with $\mathbf{T}^*$. $\mathbf{T}$ is obtained by unwrapping corresponding RGB images from randomly selected $C_\text{out}$ views using mesh output of current iteration in a differentiable way~\cite{lombardi2018deep,wei2019vr}. The resolution of $\mathbf{T}$ is 256$\times$256, and we resized $\mathbf{T}^*$ to the same resolution of $\mathbf{T}$. To normalize illumination, we perform normalized cross-correlation (NCC) on each 8$\times$8 patch of $\mathbf{T}$ and $\mathbf{T}^*$ after applying the average blur. The loss function $L_\text{tex}$ is defined as follows:
\begin{equation}
L_\text{tex} = \delta ||(\mathrm{NCC}(\mathbf{T}) - \mathrm{NCC}(\mathbf{T}^*))||_1,
\end{equation}
where $\delta$ is a binary tensor whose value is one if the corresponding UV coordinate has visible RGB value. This loss function is applied to fine-tune trained DeepHandMesh 15 epochs. The total loss function in fine-tuning stage is $L+L_\text{tex}$. During fine-tuning, we used the same learning rate $10^{-4}$, and it is reduced by a factor of 10 at the 10th and 12th epochs. $\lambda_\text{lap}$ is set to 1.

\begin{figure}[t]
\begin{center}
\includegraphics[width=1.0\linewidth]{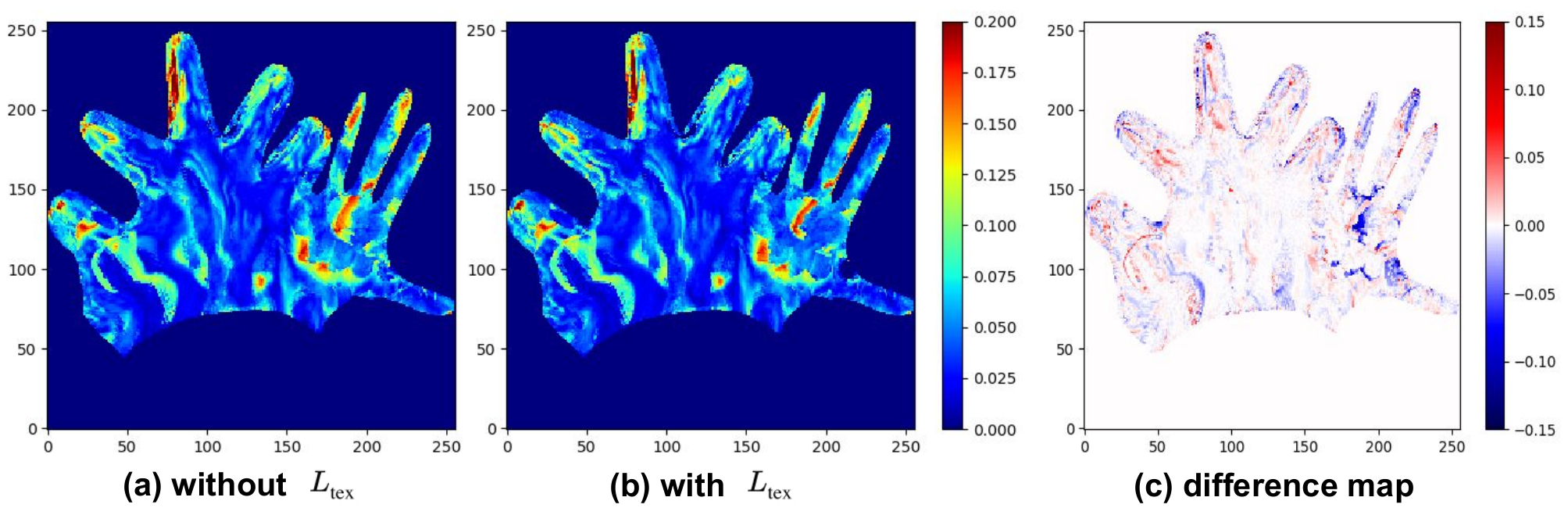}
\end{center}
\vspace*{-5mm}
   \caption{Visualization of effect of $L_{tex}$.}
\vspace*{-3mm}
\label{fig:effect_of_image_loss}
\end{figure}

Figure~\ref{fig:effect_of_image_loss} shows standard deviation $\sigma$ of each pixel value on the UV space after fine-tuning without $L_\text{tex}$ (a) and with (b). (c) shows $\sigma$ difference between (a) and (b), which is defined as (b) subtracted by (a) (\textit{i.e.}, blue colors in (c) indicate $\sigma$ decreased after fine-tuning). As the figures show, $L_\text{tex}$ helps to decrease $\sigma$, but not at a noticeable amount. We guess that this is because (a) already shows low standard deviation.

\section{Effect of the skeleton corrective}
To demonstrate the effectiveness of the identity-dependent skeleton corrective $\Delta \mathbf{S}_\beta$, we compare 3D joint distance error $\mathrm{P_{err}}$ in Table~\ref{table:effect_of_skeleton_corrective}. The error is defined as a Euclidean distance between $\mathbf{P}$ and $\mathbf{P}^*$, where $*$ indicates groundtruth. The table shows our skeleton corrective refines the skeleton of the initial hand model successfully.

\begin{table}
\centering
\setlength\tabcolsep{1.0pt}
\def\arraystretch{1.1}
\begin{tabular}{C{2.3cm}C{2.5cm}}
\specialrule{.1em}{.05em}{.05em}
Settings & $\mathrm{P_{err}}$ (mm)  \\ \hline
Without $\Delta \mathbf{S}_{\beta}$ &   4.82  \\
\textbf{Ours (full)} &  \textbf{2.38} \\ \hline
\end{tabular}
\caption{3D joint distance error $\mathrm{P_{err}}$ (mm) comparison between with and without our identity-dependent skeleton corrective $\Delta \mathbf{S}_{\beta}$.}
\vspace*{-3mm}
\label{table:effect_of_skeleton_corrective}
\end{table}

\begin{figure}[t]
\begin{center}
\includegraphics[width=0.85\linewidth]{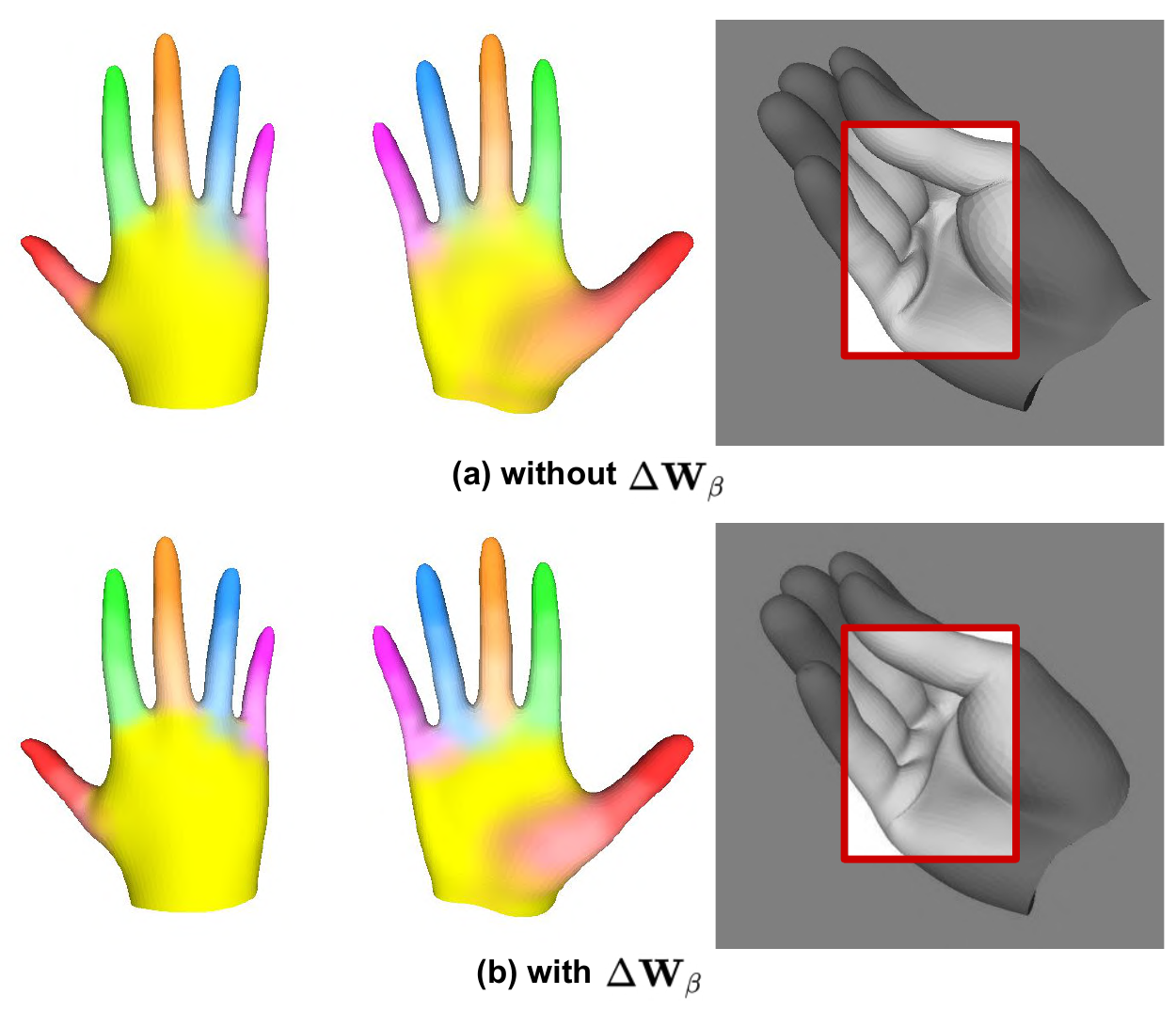}
\end{center}
\vspace*{-5mm}
   \caption{Color-coded skinning weight (left and middle) and deformed hand mesh (right) comparison between without and with skinning weight corrective $\Delta \mathbf{W}_\beta$.}
\vspace*{-3mm}
\label{fig:skinning_weight_corrrective}
\end{figure}

\section{Skinning weight corrective}
To refine pre-defined skinning weight $\mathbf{W}$, we estimate identity-dependent skinning weight corrective $\Delta \mathbf{W}_\beta \in \mathbb{R}^{V \times J}$ from the identity vector $\beta$ using two fully-connected layers. The refined skinning weight $\mathbf{W}^* \in \mathbb{R}^{V \times J}$ is obtained as follows:
\begin{equation*}
\mathbf{w}^*_{v,j} = \frac{\mathrm{max}(\mathbf{w}_{v,j} + \Delta \mathbf{w}_{v,j},0)}{\sum_{j=1}^J \mathrm{max}(\mathbf{w}_{v,j} + \Delta \mathbf{w}_{v,j},0)}, v = 1,\ldots,V, j = 1,\ldots,J,
\end{equation*}
where $\mathbf{w}^*_{v,j}$, $\mathbf{w}_{v,j}$, and $\Delta \mathbf{w}_{v,j}$ denote refined skinning weight, initial skinning weight, and skinning weight corrective of $j$th joint of $v$th vertex from $\mathbf{W}^*$, $\mathbf{W}$, and $\Delta \mathbf{W}_\beta$, respectively.
We clamp the refined skinning weight to be positive value and normalize it to make the summation 1. To encourage locality like Loper~et al.~\cite{loper2015smpl}, $\Delta \mathbf{w}_{v,j}$ is estimated only when $\mathbf{w}_{v,j} \neq 0$. Otherwise, it is set to zero.

We trained the DeepHandMesh with an additional $\Delta \mathbf{W}_\beta$. Figure~\ref{fig:skinning_weight_corrrective} shows color-coded skinning weight and deformed hand mesh. As the figure shows, the skinning weight of mainly finger root parts changed, and this change results in different skin deformation around the finger root parts. However, as there is no groundtruth mesh, it is hard to tell which hand mesh is more realistic clearly. We believe this skinning weight corrective $\Delta \mathbf{W}_\beta$ can be helpful when initial skinning weight $\mathbf{W}$ is bad and a bottleneck of better performance.

\section{Dataset description}
We provide detailed descriptions and visualizations of the sequences in our newly constructed dataset.
The pre-defined hand poses include various sign languages that are frequently used in daily life and extreme poses where each finger assumes a maximally bent or extended.
When capturing the conversational gestures, subjects are instructed with minimal instructions, for example, waving their hands as if telling someone to come over. 
The hand poses in our dataset are carefully chosen to sample a variety of poses and conversational gestures while being easy to follow by capture participants.

The hand pose estimator was trained on our held-out human annotation dataset, which includes the 3D rotation center coordinates of hand joints from 6K frames. 
The predicted 2D hand joint locations of each view were triangulated with RANSAC to robustly obtain the groundtruth 3D hand joint coordinates. 
The hand pose estimator used to obtain groundtruth 3D hand joint coordinates achieves \emph{2.78 mm error} on our held-out human-annotated test set, which is quite low. From the 3D reconstruction, we rendered groundtruth depth maps for all camera views.

\begin{figure}
\begin{center}
\includegraphics[width=1.0\linewidth]{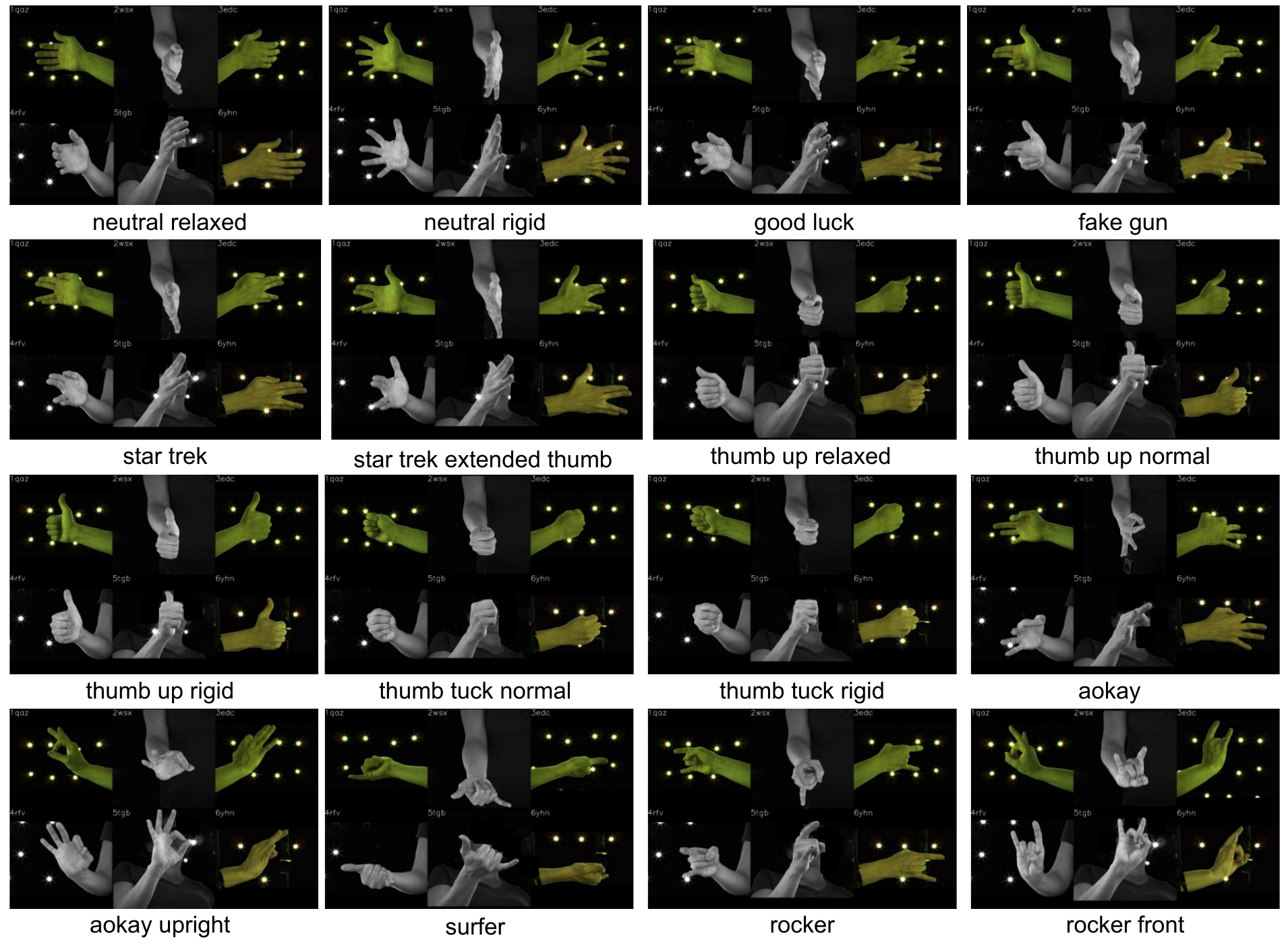}
\end{center}
\vspace*{-5mm}
   \caption{Visualization of the sequences in the training set.}
\vspace*{-3mm}
\label{fig:single_pp_1}
\end{figure}

\begin{figure}
\begin{center}
\includegraphics[width=1.0\linewidth]{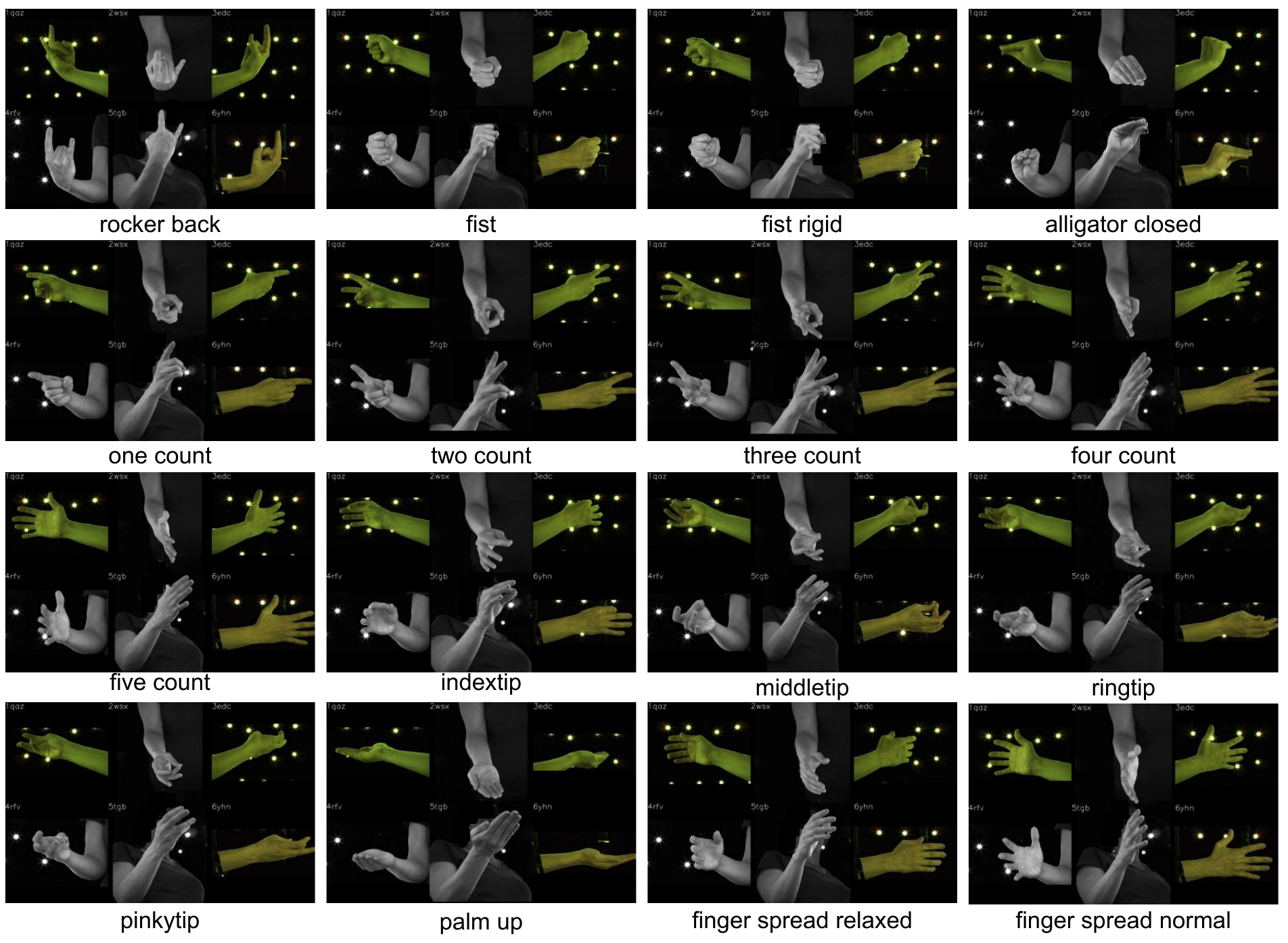}
\end{center}
\vspace*{-5mm}
   \caption{Visualization of the sequences in the training set.}
\vspace*{-3mm}
\label{fig:single_pp_2}
\end{figure}

\begin{figure}
\begin{center}
\includegraphics[width=1.0\linewidth]{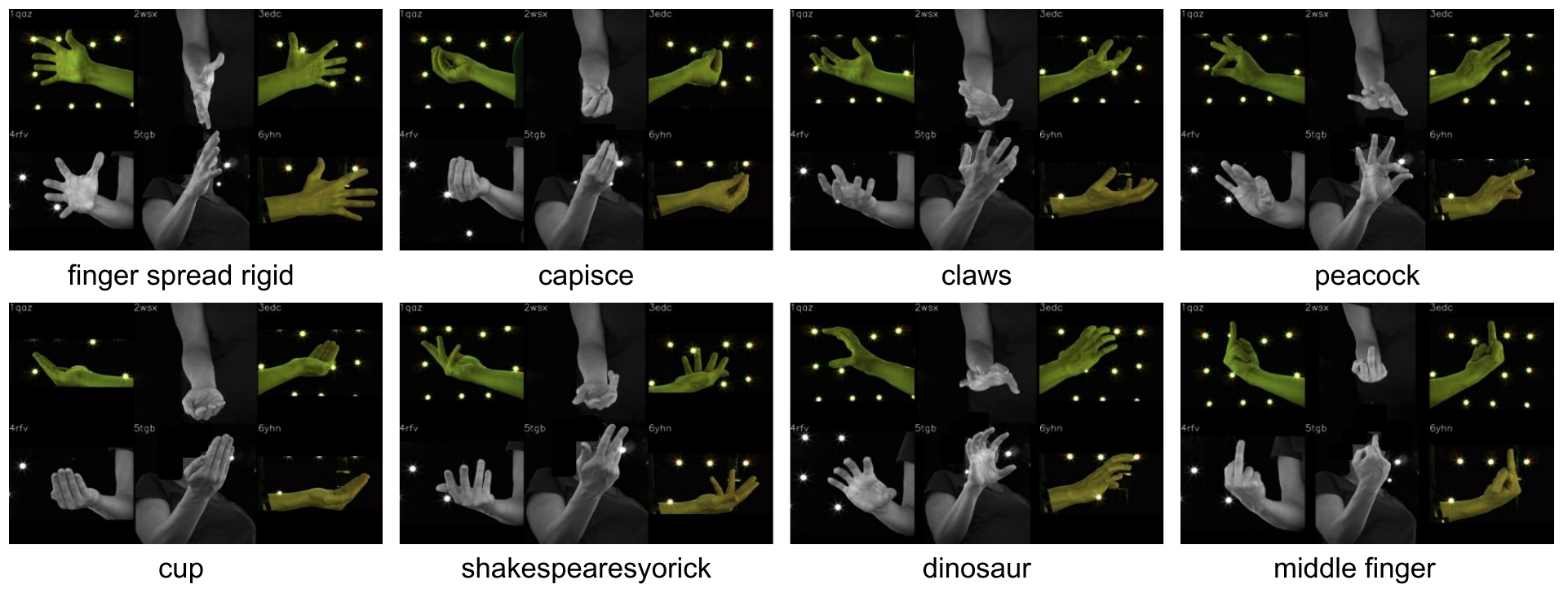}
\end{center}
\vspace*{-5mm}
   \caption{Visualization of the sequences in the training set.}
\vspace*{-3mm}
\label{fig:single_pp_3}
\end{figure}

\noindent\textbf{Training set.} 
Figure~\ref{fig:single_pp_1}, ~\ref{fig:single_pp_2}, and ~\ref{fig:single_pp_3} show the pre-defined hand poses in training set.
Belows are detailed descriptions of each sequence.
\begin{itemize}
\item neutral relaxed: the neutral hand pose. Hands in front of the chest, fingers do not touch, and palms face the side.
\item neutral rigid: the neutral hand pose with maximally extended fingers, muscles tense.
\item good luck: hand sign language with crossed index and middle fingers.
\item fake gun: hand gesture mimicking the gun.
\item star trek: hand gesture popularized by the television series Star Trek.
\item star trek extended thumb: \enquote{star trek} with extended thumb.
\item thumb up relaxed: hand sign language that means \enquote{good}, hand muscles relaxed.
\item thumb up normal: \enquote{thumb up}, hand muscles average tenseness.
\item thumb up rigid: \enquote{thumb up}, hand muscles very tense.
\item thumb tuck normal: similar to fist, but the thumb is hidden by other fingers.
\item thumb tuck rigid: \enquote{thumb tuck}, hand muscles very tense.
\item aokay: hand sign language that means \enquote{okay}, where palm faces the side.
\item aokay upright: \enquote{aokay} where palm faces the front.
\item surfer: the SHAKA sign.
\item rocker: hand gesture that represents rock and roll, where palm faces the side.
\item rocker front: the \enquote{rocker} where palm faces the front.
\item rocker back: the \enquote{rocker} where palm faces the back.
\item fist: fist hand pose.
\item fist rigid: fist with very tense hand muscles.
\item alligator closed: hand gesture mimicking the alligator with a closed mouth.
\item one count: hand sign language that represents \enquote{one.}
\item two count: hand sign language that represents \enquote{two.}
\item three count: hand sign language that represents \enquote{three.}
\item four count: hand sign language that represents \enquote{four.}
\item five count: hand sign language that represents \enquote{five.}
\item indextip: thumb and index fingertip are touching.
\item middletip: thumb and middle fingertip are touching.
\item ringtip: thumb and ring fingertip are touching.
\item pinkytip: thumb and pinky fingertip are touching.
\item palm up: has palm facing up.
\item finger spread relaxed: spread all fingers, hand muscles relaxed.
\item finger spread normal: spread all fingers, hand muscles average tenseness.
\item finger spread rigid: spread all fingers, hand muscles very tense.
\item capisce: hand sign language that represents \enquote{got it} in Italian.
\item claws: hand pose mimicking claws of animals.
\item peacock: hand pose mimicking peacock.
\item cup: hand pose mimicking a cup.
\item shakespeareyorick: hand pose from Yorick from Shakespeare's play Hamlet.
\item dinosaur: hand pose mimicking a dinosaur.
\item middle finger: hand sign language that has an offensive meaning.
\end{itemize}

\clearpage

\begin{figure}
\begin{center}
\includegraphics[width=1.0\linewidth]{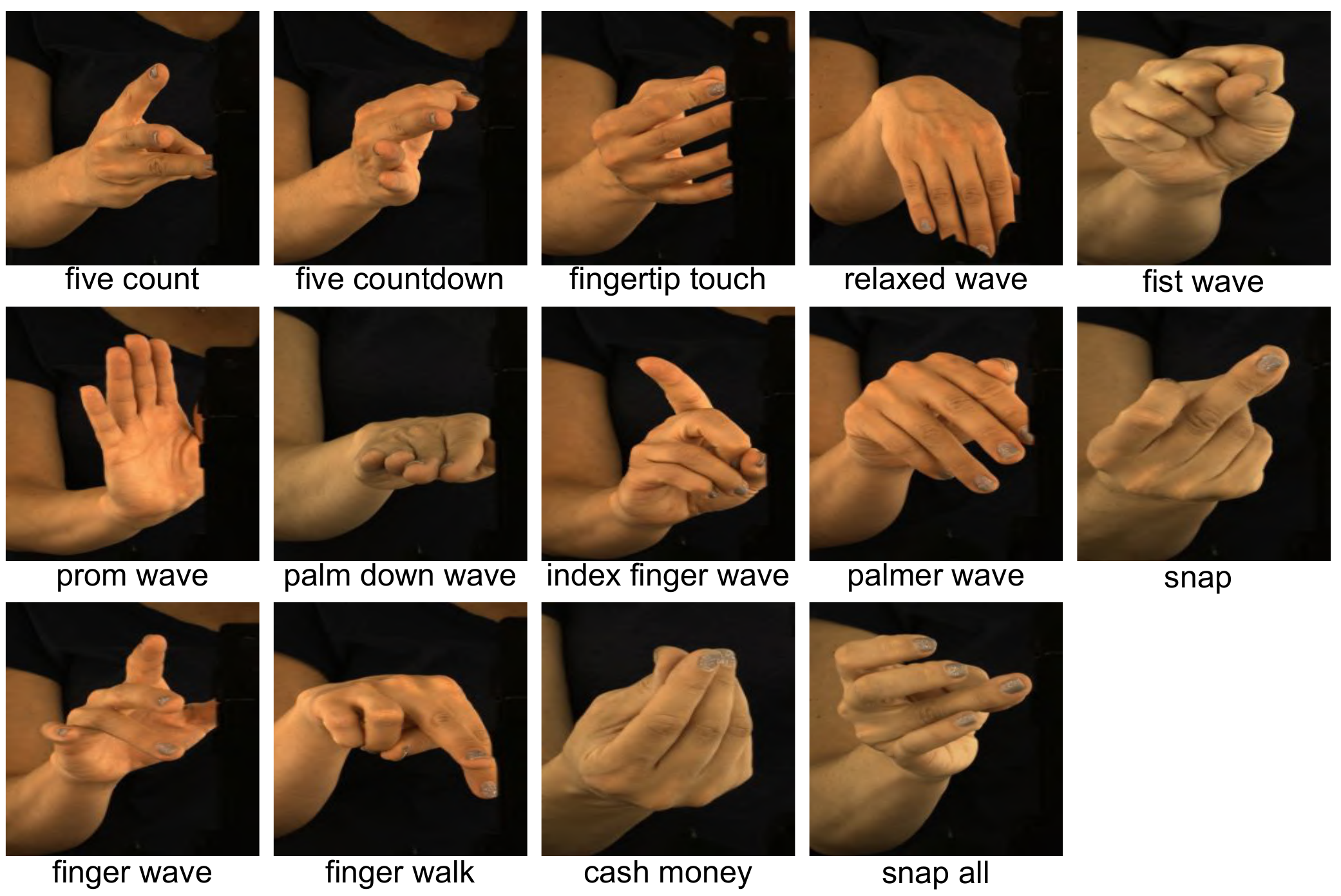}
\end{center}
\vspace*{-5mm}
   \caption{Visualization of the sequences in the testing set.}
\vspace*{-3mm}
\label{fig:single_rom}
\end{figure}

\noindent\textbf{Testing set.} 
Figure~\ref{fig:single_rom} shows the conversational gestures in testing set. 
Belows are detailed descriptions of each sequence.
\begin{itemize}
\item five count: count from one to five. 
\item five countdown: count from five to one.
\item fingertip touch: thumb touch each fingertip.
\item relaxed wave: wrist relaxed, fingertips facing down and relaxed, wave.
\item fist wave: rotate wrist while hand in a fist shape.
\item prom wave: wave with fingers together.
\item palm down wave: wave hand with the palm facing down.
\item index finger wave: hand gesture that represents \enquote{no} sign.
\item palmer wave: palm down, scoop towards you, like petting an animal.
\item snap: snap middle finger and thumb.
\item finger wave: palm down, move fingers like playing the piano.
\item finger walk: mimicking a walking person by index and middle finger.
\item cash money: rub thumb on the index and middle fingertips.
\item snap all: snap each finger on the thumb.
\end{itemize}

\newpage

\begin{figure}
\begin{center}
\includegraphics[width=1.0\linewidth]{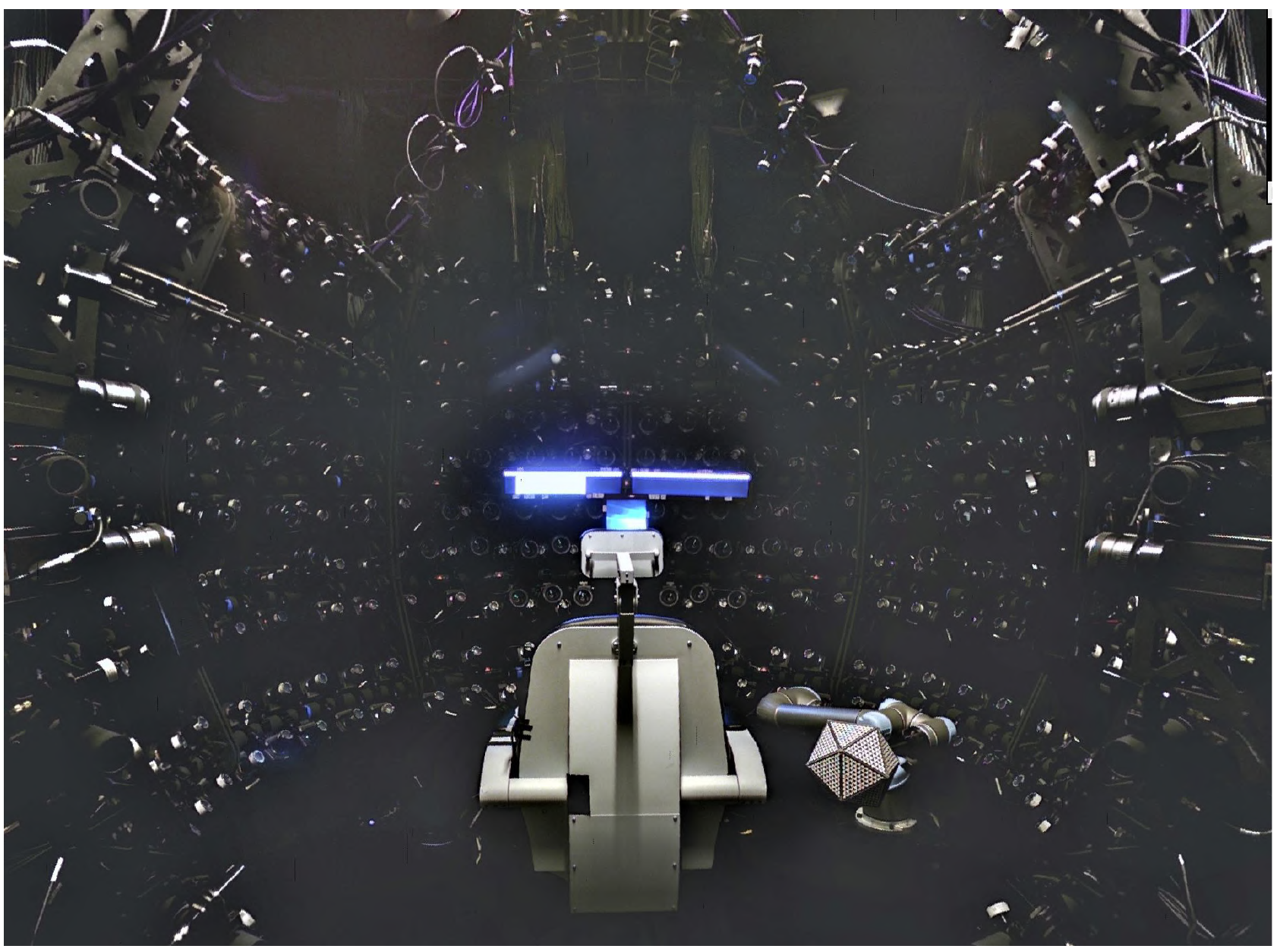}
\end{center}
\vspace*{-5mm}
   \caption{Rendering of our constructed multi-view studio.}
\vspace*{-3mm}
\label{fig:studio}
\end{figure}

Figure~\ref{fig:studio} shows rendering of our constructed multi-view studio for the data capture.

\clearpage

\section{Comparison with state-of-the-art methods}
\begin{figure}[t]
\begin{center}
\includegraphics[width=0.7\linewidth]{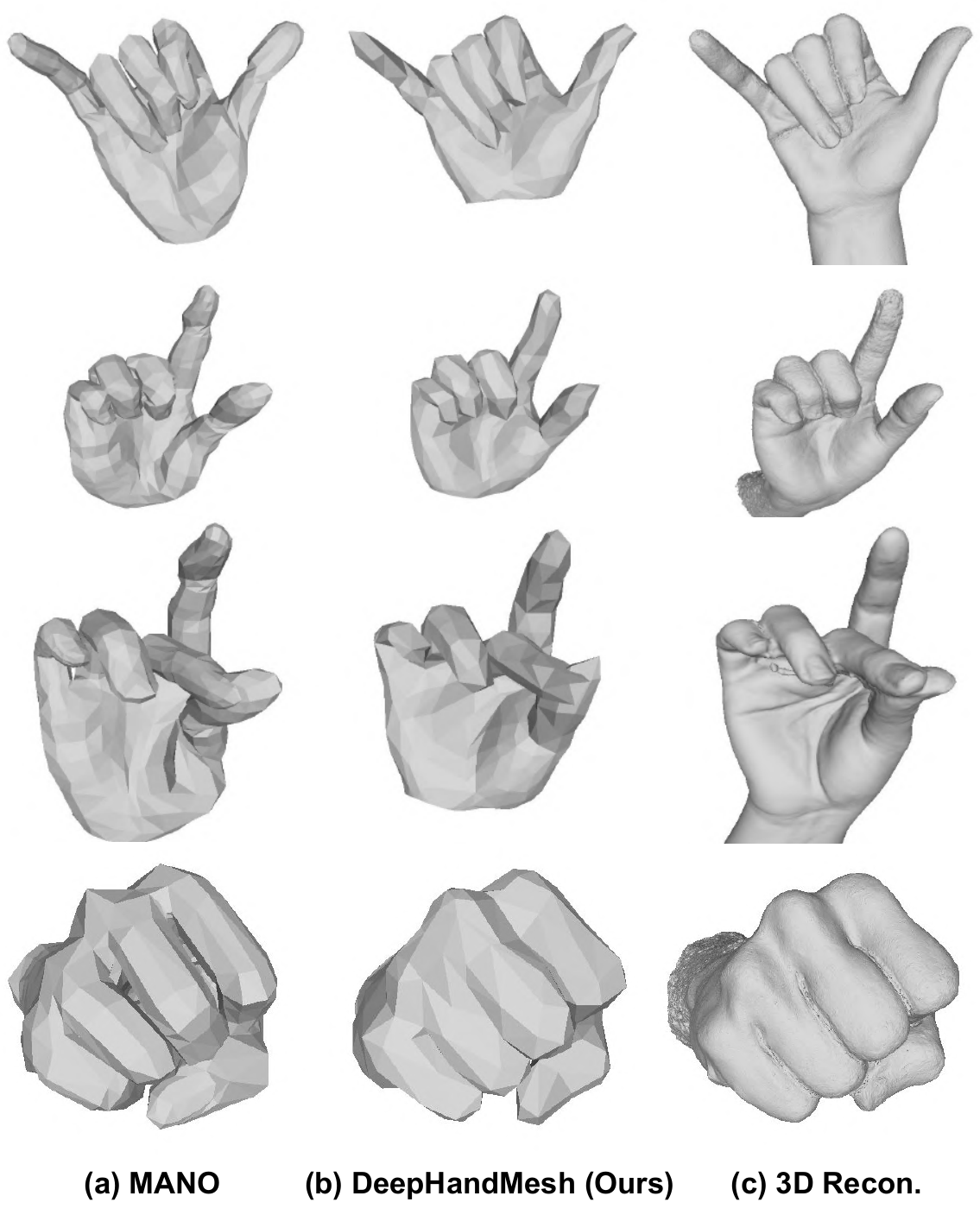}
\end{center}
\vspace*{-7mm}
   \caption{Estimated hand mesh comparison with MANO~\cite{romero2017embodied} and our DeepHandMesh using the hand model of the similar resolution.}
\vspace*{-5mm}
\label{fig:comparison_with_mano_lr}
\end{figure}

\noindent\textbf{Comparison with MANO under the similar mesh resolution.}
We train and test our DeepHandMesh with a hand model the resolution of which is similar to that of MANO, and compare its qualitative results with those from MANO in Figure~\ref{fig:comparison_with_mano_lr}.
Our low-resolution DeepHandMesh uses a hand mesh model with 792 vertices, while the MANO is based on a hand mesh model with 778 vertices. 
The figure shows that our DeepHandMesh provides a more realistic hand mesh compared with MANO under the similar resolution of the hand mesh model. Note that our DeepHandMesh is trained in a weakly-supervised way without per-vertex loss function, while MANO is based on fully-supervised training with per-vertex loss. When we train low resolution version of the DeepHandMesh, $L2$ norm regularizers are used for the correctives (\textit{i.e.}, $\Delta \mathbf{S}_\beta$, $\Delta \mathbf{M}_\beta$, and $\Delta \mathbf{M}_\theta$).

\noindent\textbf{Comparison with MANO on the dataset of MANO.}
We tried to train and test our DeepHandMesh on MANO dataset~\cite{romero2017embodied}. However, we observed that there are only 50 registrations available for each subject, 
which are not large enough to train DeepHandMesh. 
Also, some of the 3D scans include an object grasped by a hand. 
This makes training our system on the MANO dataset hard because rendered groundtruth depth maps $\mathcal{D}^*$ include those objects. 
Although we tried to use the registered meshes that do not include the objects for depth map rendering, we noticed that the rendered depth map lost high-frequency information in the original 3D scans because of the low-resolution mesh in MANO,
which makes the depth maps hard to be used as groundtruth depth maps $\mathcal{D}^*$.

\begin{figure}[t]
\begin{center}
\includegraphics[width=0.4\linewidth]{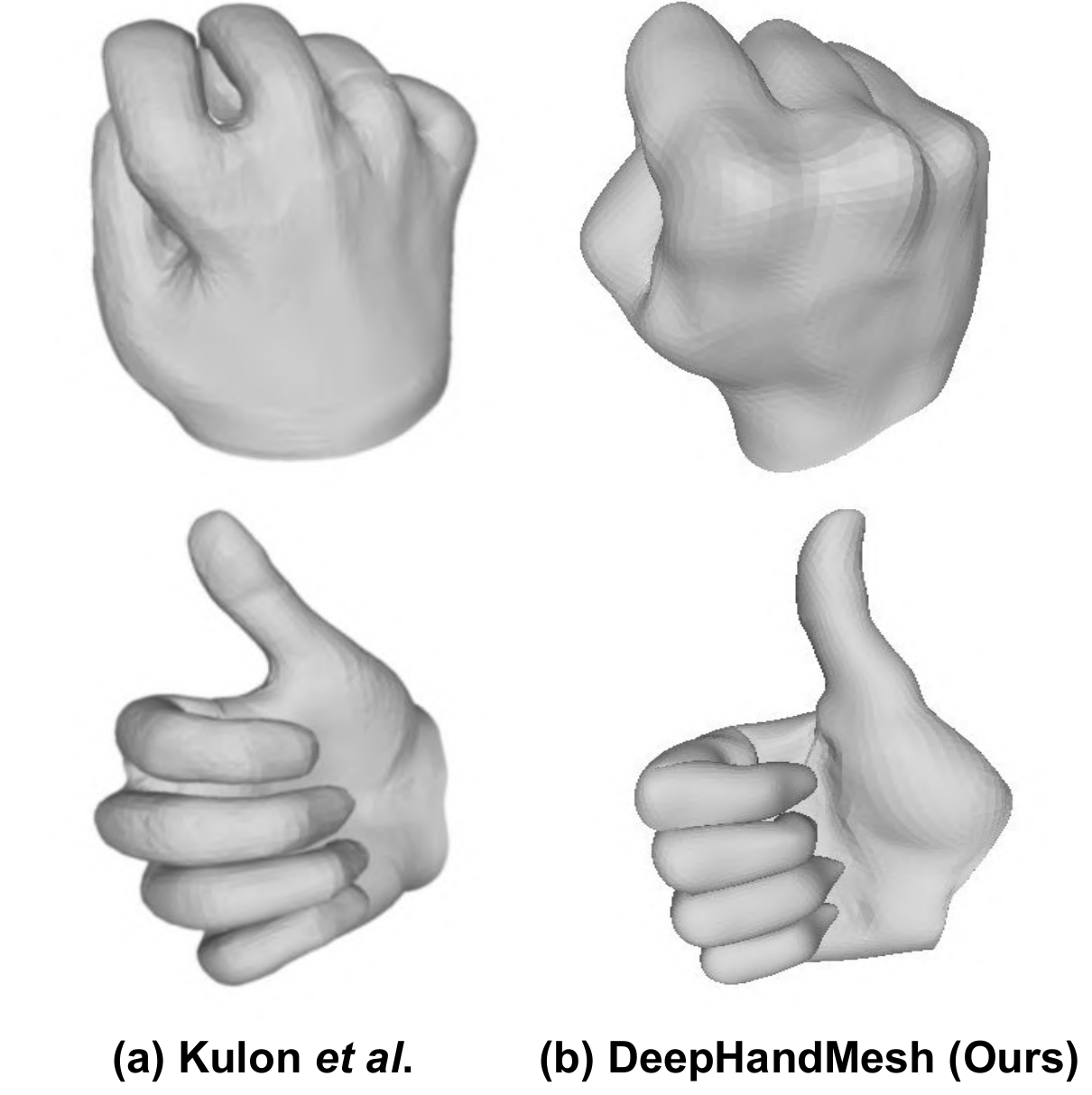}
\end{center}
\vspace*{-7mm}
   \caption{Estimated hand mesh comparison with Kulon~et al.~\cite{kulon2019single} and our DeepHandMesh. The results of Kulon~et al.~\cite{kulon2019single} are taken from their paper.}
\vspace*{-5mm}
\label{fig:comparison_with_kulon}
\end{figure}

\noindent\textbf{Comparison with Kulon~et al.~\cite{kulon2019single}.}
We also tried to compare our DeepHandMesh with Kulon~et al.~\cite{kulon2019single}. They use mesh supervision when they train their high-resolution hand model (\textit{i.e}, 7,907 vertices). As there is no mesh groundtruth in our dataset, we trained their model with the same loss functions as ours (\textit{i.e.}, \textbf{Pose loss} and \textbf{Depth map loss}). We observed that without per-vertex mesh supervision, their model provides severely distorted hand mesh. We could not train our DeepHandMesh on the Panoptic dome dataset~\cite{simon2017hand} that Kulon~et al.~\cite{kulon2019single} used because high-quality multi-view depth maps are not available in that dataset. Instead, we compare hand mesh output of the same hand pose from our DeepHandMesh and Kulon~et al.~\cite{kulon2019single} trained on our dataset and the Panoptic dome dataset~\cite{simon2017hand}, respectively. Although this is not a perfectly fair comparison, we think that this can roughly show how the final outputs of each method are different. Figure~\ref{fig:comparison_with_kulon} shows our DeepHandMesh provides significantly more realistic hand mesh than Kulon~et al.~\cite{kulon2019single}.

\noindent\textbf{Comparison on publicly available datasets.}
As our DeepHandMesh is a personalized system, it is hard to compare with other hand models (\textit{i.e.}, MANO~\cite{romero2017embodied} and Kulon~et al.~\cite{kulon2019single}) that support cross-identity on publicly available datasets.
Instead, we tried to best to provide comparisons between them on our dataset.

\clearpage
 
\clearpage

\section{Qualitative rendered results}
\begin{figure}
\begin{center}
\includegraphics[width=0.8\linewidth]{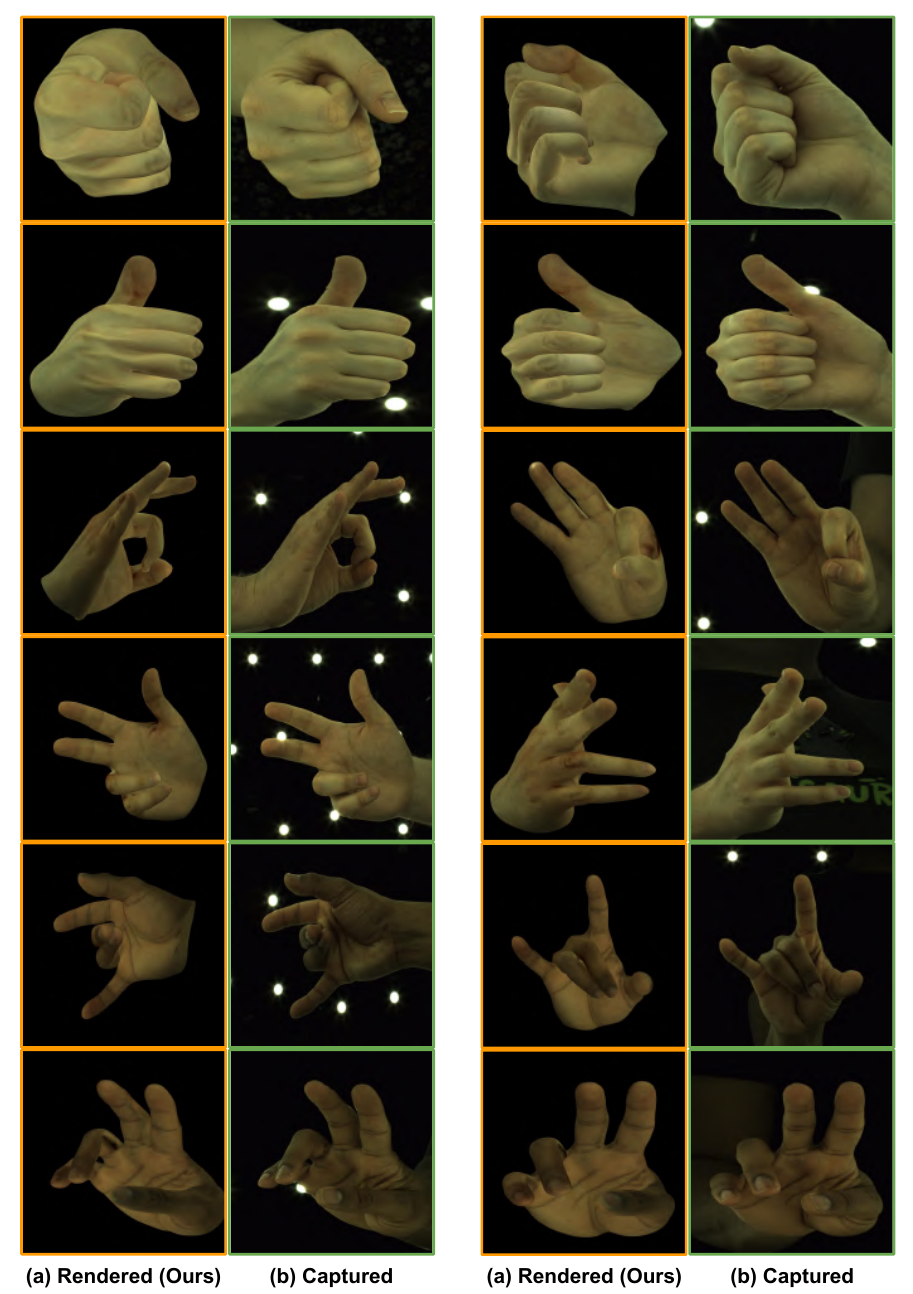}
\end{center}
\vspace*{-5mm}
   \caption{Comparison between our rendered image and captured image from cameras.}
\vspace*{-3mm}
\label{fig:render_result}
\end{figure}

We provide rendered result using texture obtained from Section~\ref{sec:texture} in Figure~\ref{fig:render_result}.

\clearpage

%
%
\bibliographystyle{splncs04}
\bibliography{main}
\end{document}